\documentclass[journal,twoside,web]{ieeecolor}
\usepackage{generic}
\usepackage{cite}
\usepackage{amsmath,amssymb,amsfonts}
\usepackage{algorithmic}
\usepackage{graphicx}
\usepackage{algorithm,algorithmic}
\usepackage{hyperref}
\hypersetup{hidelinks=true}
\usepackage{textcomp}
\usepackage{xcolor}
\usepackage{booktabs}
\usepackage{float}
\usepackage[normalem]{ulem}
\usepackage{colortbl}

\usepackage{multirow}

\def\BibTeX{{\rm B\kern-.05em{\sc i\kern-.025em b}\kern-.08em
    T\kern-.1667em\lower.7ex\hbox{E}\kern-.125emX}}
\begin{document}
\title{PBE-UNet: A light weight Progressive Boundary-Enhanced U-Net with Scale-Aware Aggregation for Ultrasound Image Segmentation}
\author{Chen Wang, Yixin Zhu, Yongbin Zhu, Fengyuan Shi, Qi Li, Jun Wang, Zuozhu Liu, Keli Hu
\thanks{This work was supported in part by the National Natural Science
Foundation of China under Grant 62476237, the Humanities and Social Sciences Youth Foundation of Ministry of Education of China under Grant 24YJCZH283 and 25YJCZH082, and the Zhejiang Provincial Natural Science Foundation of China under Grant LZ24F020006. (Corresponding author: Keli Hu). }
\thanks{Chen Wang, Yinxin Zhu, and Yongbin Zhu are with the Department of Computer Science, Shaoxing University, Shaoxing, China (e-mail: wangchen3024@163.com; m17858908309@163.com; 24020855130@usx.edu.cn).}
\thanks{Fengyuan Shi is with the National Frontiers Science Center for Industrial Intelligence and Systems Optimization, Northeastern University, China(e-mail: im.fengyuan.shi@outlook.com).}
\thanks{Qi Li is with the College of Computer Science, Chongqing University, China(e-mail: qili@cqu.edu.cn).}
\thanks{Jun Wang is with the School of Computer and Computing Science, Hangzhou City University, Hangzhou, China(e-mail: wangjun@hzcu.edu.cn). }
\thanks{Zuozhu Liu is with the ZJU-UIUC Institute, Zhejiang University, Haining, China (e-mail: zuozhuliu@intl.zju.edu.cn). }
\thanks{Keli Hu is with the Department of Computer Science, Shaoxing University, Shaoxing, China, and also with the Information Technology R\&D Innovation Center of Peking University, Shaoxing, China (e-mail: ancimoon@gmail.com).}
}

\maketitle
\begin{abstract}

Accurate lesion segmentation in ultrasound images is essential for preventive screening and clinical diagnosis, yet remains challenging due to low contrast, blurry boundaries, and significant scale variations. Although existing deep learning-based methods have achieved remarkable performance, these methods still struggle with scale variations and indistinct tumor boundaries. To address these challenges, we propose a progressive boundary enhanced U-Net (PBE-UNet). Specially, we first introduce a scale-aware aggregation module (SAAM) that dynamically adjusts its receptive field to capture robust multi-scale contextual information. 
Then, we propose a boundary-guided feature enhancement (BGFE) module to enhance the feature representations. We find that there are large gaps between the narrow boundary and the wide segmentation error areas. Unlike existing methods that treat boundaries as static masks, the BGFE module progressively expands the narrow boundary prediction into broader spatial attention maps. Thus, broader spatial attention maps could effectively cover the wider segmentation error regions and enhance the model's focus on these challenging areas. We conduct expensive experiments on four benchmark ultrasound datasets, BUSI, Dataset B, TN3K, and BP. The experimental results how that our proposed PBE-UNet outperforms state-of-the-art ultrasound image segmentation methods. The code is at https://github.com/cruelMouth/PBE-UNet.

\end{abstract}


\section{Introduction}
\label{sec:introduction}
\IEEEPARstart{U}{ltrasound} image has emerged as a primary radiological modality for the screening of breast tumors, thyroid nodules, and various other clinical conditions, due to its non-invasive nature, cost-effectiveness, and real-time feedback~\cite{masnetbibm2025}. Accurate segmentation of the lesion is an indispensable prerequisite for computer-aided clinical diagnosis~\cite{qin2025mbe}. However, the inherent characteristics of ultrasound images, such as low contrast, blurred boundaries, and the varying shapes and sizes of targets, make it difficult to precisely distinguish lesions from surrounding tissues, severely impacting the performance of target segmentation~\cite{xue2021global}.

\begin{figure}[!t]
	\centering  
	\includegraphics[width=\linewidth]{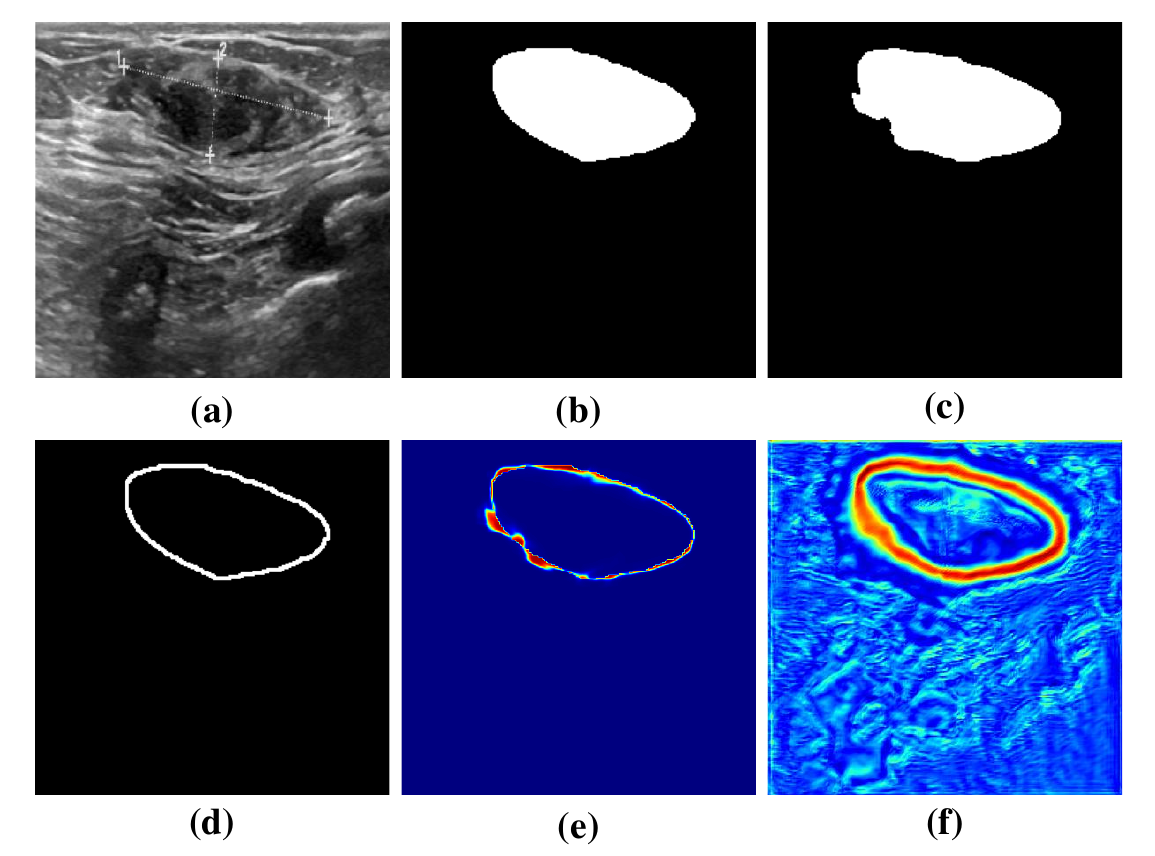} 
	\caption{Visualization of prediction map, GT boundary, error map, and the boundary attention map generated by our method. (a) original image, (b) GT, (c) prediction map of U-Net, (d) GT boundary, (e) error map, (f) boundary attention map generated by our method. We can find that the narrow boundary fails to cover the wider error region, highlighting the limitation of simple feature fusion methods, such as ``Add'', ``Multiply'', and ``Contact''. In contrast, our BGFE module addresses this issue by adaptively expanding the boundary into a broader attention region, which effectively aligns with and covers the error areas.} 
	\label{fig:picBlur}  
\end{figure}
Recently, deep learning-based segmentation methods have achieved considerable progress in medical image segmentation.
Convolutional neural networks(CNN) architectures, such as U-Net~\cite{ronneberger2015u}, U-Net++~\cite{zhou2019unet++} and Attention U-Net~\cite{oktay2018attention}, have achieved remarkable success. However, these CNN-based methods usually struggle to model long-range dependencies due to their fixed receptive fields. To mitigate these issues, researchers have introduced the vision transformer and proposed various CNN-Transformer hybrid medical image segmentation methods~\cite{chen2021transunet}. These hybrid methods take the advantage of CNN's inductive bias and the vision Transformer's global context extraction capabilities, thereby effectively capturing both local details and global context. For example, Wang et al.~\cite{masnetbibm2025} employs the PVTv2 as the backbone and propose a foreground and background separable encoding architecture for ultrasound image segmentation. Furthermore, significant variation in lesion size, shape, and location demands robust multi-scale feature representation. Several multi-scale feature fusion strategies~\cite{chen2022aau,tang2023cmu,chen2024esknet,wang2025pconv} have been proposed. These methods employ multi-scale dilated or depth-wise convolutions to extract and aggregate features at different scales. Although these methods have achieved substantial progress, these methods still exhibit notable computational overhead and limited adaptability in dynamically adjusting receptive fields. It still deserves further exploration toward efficient and scale-robust ultrasound image segmentation.


Leveraging boundary information to guide segmentation has proven effective for addressing ambiguous boundaries in medical image segmentation~\cite{ZhaoLFCYC19, zhou2025edge}. Early approaches~\cite{kervadec2021boundary, xue2021global} often employ handcraft operators such as Sobel and Canny, to extract boundaries and treat boundary prediction as an auxiliary supervision task. Subsequent studies~\cite{lin2025rethinking} attempt to explore the boundary-assisted feature learning through operations such as addition~\cite{ZhouRL_MIA_26}, element-wise multiplication~\cite{BGRA-Net21, BGRA-GSA23}, and channel concatenation~\cite{du2022icgnet, qin2025mbe}. However, as illustrated in Fig.~\ref{fig:picBlur}, these methods share a fundamental limitation. These methods typically employ the predicted boundary—a thin, single-pixel-wide curve—as a static mask to enhance the features. In ultrasound images, segmentation errors often spread across a broader region around the true boundary. This leads to a spatial mismatch between the ``narrow boundary'' and the ``wider error band''. Consequently, the boundary prior information is underutilized and ineffectively applied. This limitation ultimately restricts further performance improvement.


To overcome the aforementioned limitations, we propose PBE-UNet, a Progressive Boundary Enhanced UNet designed for precise and efficient ultrasound image segmentation. Recognizing the limitations of static multi-scale filters, we first integrate a scale-aware aggregation module (SAAM) at each decoder stage. This module dynamically adapts its receptive field to capture multi-scale contextual information, ensuring robust representation across lesions of drastically different sizes. To resolve the spatial mismatch between thin boundary priors and wider error regions, we introduce a dual-stage boundary refinement strategy. Specifically, the boundary detection (BD) module is employed to generate initial boundary map. This boundary functions as explicit supervisory information, forming a spatial prior that is subsequently refined by the boundary-guided feature enhancement (BGFE) module. Unlike traditional methods that treat boundaries as static masks, the BGFE module progressively expands and refines boundary semantics, effectively bridging the gap between narrow priors and the actual error-prone areas. By leveraging these refined features as a spatial attention mechanism, PBE-UNet maintains an explicit focus on ambiguous regions, ultimately achieving superior detail recovery.


The main contributions of this paper are summarized as follows:
\begin{itemize}
    \item We present a novel progressive boundary-enhanced UNet architecture, named PBE-UNet. It effectively integrates boundary guided feature enhancement with multi-scale feature aggregation. This integration significantly enhance the segmentation accuracy of ultrasound lesions, particularly for targets with ambiguous boundaries and large size variations.
    \item We propose an innovative Boundary-Guided Feature Enhancement (BGFE) module to address the ambiguous boundaries. This module adaptively expands the narrow boundary prediction into broader spatial attention maps, effectively covering the wider segmentation error regions. Consequently, it enhances the model's focus on these challenging areas.
    \item We introduce a light weight Scale-Aware Aggregation Module (SAAM) to tackle the inherent challenge of lesion scale variations. This module leverages parallel depthwise dilated convolutions with varying receptive fields to adaptively captures multi-scale context. It provides robust feature representations for the decoder with low computational overhead.
    \item We conduct extensive experiments on four benchmark ultrasound datasets: BUSI, Dataset B, TN3K, and BP. The experimental results demonstrate that our proposed PBE-UNet consistently outperforms several state-of-the-art ultrasound image segmentation methods.

\end{itemize}

\section{Related Work}

\subsection{Ultrasound Image Segmentation}

Advances in ultrasound image segmentation have closely followed the progression of deep learning architectures, moving from CNNs-enhanced architectures~\cite{chen2022aau,chen2024esknet} to hybrid transformer-based designs~\cite{qu2024EH-former,liu2024cross,msanet2025}. CNN-based methods often focus on stronger local feature extraction. For example, He et al.~\cite{he2020dense} propose to employ dense biased connections to compress and forward feature maps across subsequent layers. Ning et al.\cite{ning2021smu} propose a saliency-guided morphology-aware network, which learns and fuses salient foreground and background features to boost overall model performance. Yin et al.~\cite{CFU-Net2023} propose a coarse-to-fine framework, generating coarse predictions and then employing them to refine the final predictions. Despite these advances, CNNs-based methods remain limited in capturing long-range dependencies. To address this, many Transformer-CNN hybrid models have been designed and these hybrid methods have shown great potential in medical image segmentation. Similarity, Liu et al.~\cite{liu2024cross} introduce a parallel encoder framework, which contains a CNN branch and a transformer branch, to capture both local and global context. Qu et al.~\cite{qu2024EH-former} employ the PVTv2 as the backbone and introduce the Regional Easy-Hard-Aware Transformer architecture for hard region mining in ultrasound images. MSA-Net~\cite{msanet2025} proposes a foreground-background decoupling architecture based on separable attention for breast tumor segmentation.

The significant scale variation of lesions is a persistent challenge in ultrasound image segmentation. To capture multi-scale context, a variety of multi-scale feature aggregation strategies have been proposed. These methods usually enhance feature maps through extracting and aggregating multi-scale features within the encoder, decoders, and the skip-connection. For example, MultiResUNet~\cite{multiresunet} introduces a learnable multi-scale convolution module in the encoder stage to aggregate multi-scale features and improve the perception of irregular structures. Subsequently, AAU-Net~\cite{chen2022aau} utilizes multi-scale convolutions coupled with spatial attention to enlarge the receptive field, while CMUNet~\cite{tang2023cmu} aggregates multi-scale features at skip-connections via multi-scale dilated convolutions. ESKNet~\cite{chen2024esknet} adaptively fuses multi-branch features with an channel-spatial attention mechanism. PConv-UNet~\cite{wang2025pconv} dynamically aggregates multi-scale directional features through stacked multi-scale Pinwheel kernels and spatial attention.

Despite their effectiveness, many existing multi-scale modules have notable computational overhead, limiting their practicality in efficiency-sensitive scenarios. This motivates our design of the lightweight Scale-Aware Aggregation Module (SAAM). It aims to build a robust multi-scale feature foundation with low computational cost, thereby providing rich context for subsequent boundary refinement.

\subsection{Boundary-Assisted segmentation}
Some previous works~\cite{ZhaoLFCYC19,zhou2025edge} have demonstrated that boundary information is beneficial for boosting the the segmentation and detection performance in medical images. Early methods~\cite{kervadec2021boundary} usually employ the boundary detection as an auxiliary supervision signal to regularize the segmentation training process. For example, Xue et al.~\cite{xue2021global} propose to extract boundaries from encoders and provide deep supervision signals at various stages, to guide boundary region segmentation. Chen et al.~\cite{chen2023deep} utilize simple convolution operations to obtain the boundary as the supervise signals. Sun et al.~\cite{sun2023boundary} introduced a boundary difference over union loss to guide boundary region segmentation. Song et al.~\cite{song2024boundary} utilize the Laplacian operator for boundary extraction and design a weighted binary cross-entropy boundary loss to enhance segmentation performance. These approaches effectively steer the network’s focus toward edge regions.

Strategies for utilizing boundaries have shifted from employing boundary as deep supervision toward more advanced boundary-assisted feature integration. The fusion strategies can be categorized by the core operations. \textbf{Addition/Multiplication:} A straightforward approach is to add~\cite{ZhouRL_MIA_26} or element-wise multiply~\cite{BGRA-Net21,BGRA-GSA23,zhou2025edge,lin2025rethinking} the boundary maps with features. For instance, ~\cite{BGRA-Net21,BGRA-GSA23} propose to multiply the boundary maps from encoder features with the decoder features, while Zhou et al.~\cite{zhou2025edge} and Lin et al.~\cite{lin2025rethinking} propose to fuse the edge features with high-level semantic features through element-wise multiplication. \textbf{Concatenation:} Some works propose to concatenate boundary maps with features for subsequent processing. Du et al.~\cite{du2022icgnet} and Qin et al.~\cite{qin2025mbe}  enhance the features from encoders (and decoders) with the extracted boundary. Yue et al.~\cite{yue2025boundary} further employ the boundary to guide the alignment and fusion of encoders and decoders. Despite these advancements, a common characteristic persists: the extracted boundary-often a thin, single-pixel curve-is used directly as a static mask for fusion.

We argue that these fusion strategies underutilize the boundary prior. Through visual analysis (see Fig.~\ref{fig:picBlur}), we observe a critical spatial mismatch:the segmentation errors often form a wider band around the boundary, while the predicted boundary is merely a narrow curve. Consequently, direct fusion with this narrow boundary inherently limits its spatial influence and fails to cover the broader error regions. This observation motivate our core design: instead of employing the raw boundary directly, we adaptively expand the boundary into broader, spatially-aware attention maps. This expanded prior can better align with the actual error distribution, providing more comprehensive guidance for feature enhancement.

\begin{figure}[!t]
	\centering  
	\includegraphics[width=0.95\linewidth]{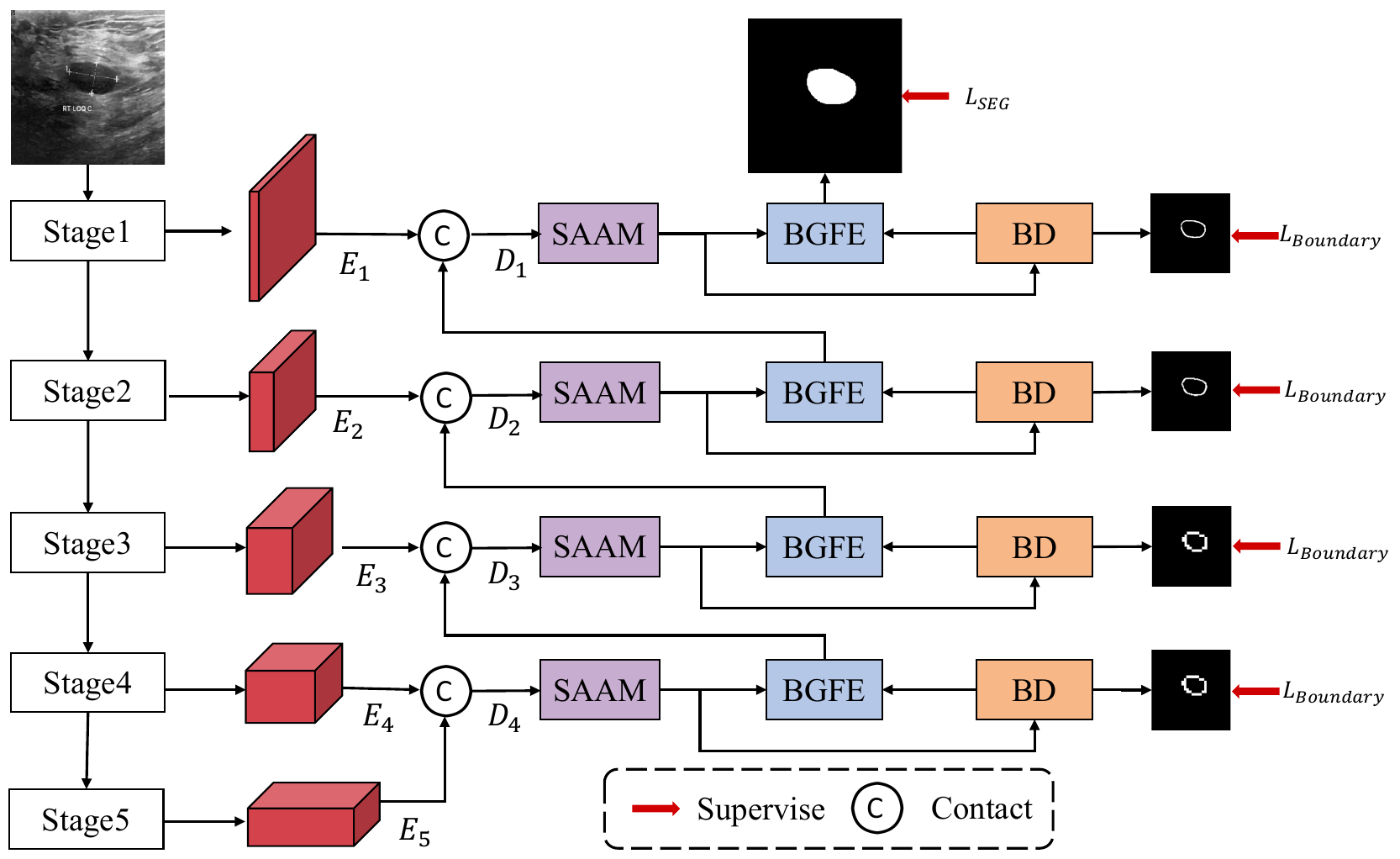} 
	\caption{The overall framework of the PBE-UNet network. It consists of the encoder, decoder, boundary detection (BD) module, boundary-guided feature enhancement (BGFE) module, and scale-aware aggregation module (SAAM). } 
	\label{fig:picOverall}  
\end{figure}

\section{METHODOLOGY}
\subsection{Overall Architecture}

The network architecture of the proposed PBE-Net is shown in Fig.~\ref{fig:picOverall}. The encoder-decoder network takes an ultrasound image as the input and predicts a mask as the output. The proposed architecture employs multiple scale-aware aggregation modules (SAAM) to aggregate multi-level features in a bottom-up manner, facilitating the extraction of robust multi-scale tumor characteristics. To fully leverage boundary information for feature enhancement, we first integrate a boundary detection (BD) module as an auxiliary task at each decoder stage. Within the boundary-guided feature enhancement (BGFE) module, our approach fuses features with predicted boundary information to mitigate noise resulting from spatial misalignment. We then refine the boundary details and adaptively enhance the input features by incorporating spatial attention mechanisms. These enhanced features eventually serve as primary inputs to the SAAM to facilitate further feature optimization.

\subsection{Boundary Detection Module }
\begin{figure}[!t]
	\centering  
	\includegraphics[width=0.95\linewidth]{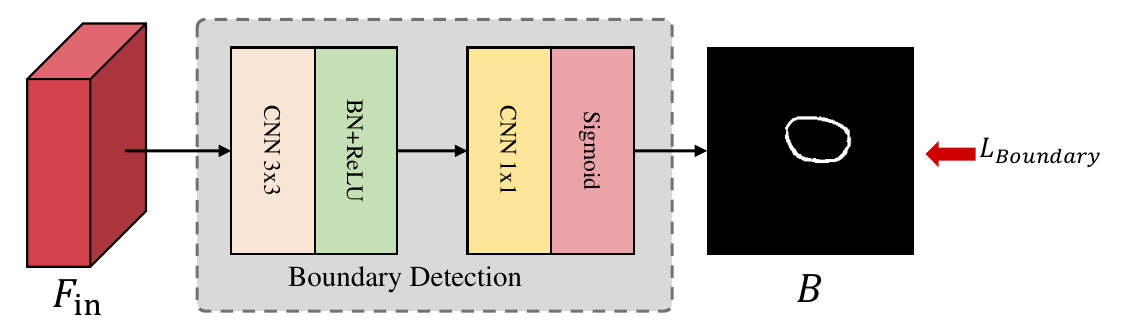} 
    \caption{The architecture of boundary detection module. } 
	\label{fig:picBD}  
\end{figure}

Different from previous methods that employ traditional edge detection operators, such as Canny, Sobel, and Laplacian, we introduce a boundary detection module to extract the boundary probability map from the input features, as shown in Fig.~\ref{fig:picBD}. The boundary detection consists of two convolution layers and two activations, as $\left[ 3\times 3, \mathrm{BN}, \mathrm{ReLU}, 1\times 1, \mathrm{Sigmoid} \right]$.
The whole process of obtaining boundary can be expressed as follows:
\begin{equation}
F'_{in}=ReLU(BN(Conv3\times3(F_{in}))),
\label{eqa:eqa1}
\end{equation}
\begin{equation}
B=Sigmiod(Conv1\times1(F'_{in})),
\label{eqa:eqa1-1}
\end{equation}
where $F_{in}$ means the input features at each stage, $F'_{in}$ is the intermediate features, $B$ indicates the predicted boundary. The boundary detection task at each stage could provide extra supervised information and boost the model paying more attention on the ambiguous regions. We find that such a simple auxiliary task could bring slight performance improvements. In the latter section, we will present how to make full use of the boundary signals to achieve more accurate breast tumor segmentation.

\subsection{Boundary-Guided Feature Enhancement Module}
Current boundary-assisted segmentation methods typically integrate boundary maps with feature maps through addition, multiplication, or channel-wise concatenation. However, our visual analysis of prediction errors reveals a spatial discrepancy: the boundary manifests as a narrow, continuous curve, whereas error regions, while adjacent, often span a broader area. Consequently, simple fusion operations are inadequate for modeling the actual error distribution, resulting in limited performance gains.

Motivated by this insight, we propose a new strategy. We use the predicted boundary as an anchor to adaptively expand a wider attention region. This expanded region could cover the broader potential error regions around the boundary. 
To implement this idea, we propose the Boundary-Guided Feature Enhancement (BGFE) module, as shown in Fig.~\ref{fig:picBGFE}. The proposed BGFE module employs the predicted boundary as spatial priors to refine the features via spatial attention. We first concatenate the predicted boundary map $B \in \mathbb{R}^{H\times W\times 1}$ and the input feature map $F_{in} \in \mathbb{R}^{H\times W\times C}$ along the channel dimension. Then we put the concatenated features $[F_{in};B]\in \mathbb{R}^{H\times W \times(C+1)}$ into the $3\times 3$ to get the fused representation $F_{fused}\in \mathbb{R}^{H\times W \times C}$ as follows:
\begin{figure}[!t]
	\centering  
	\includegraphics[width=0.95\linewidth]{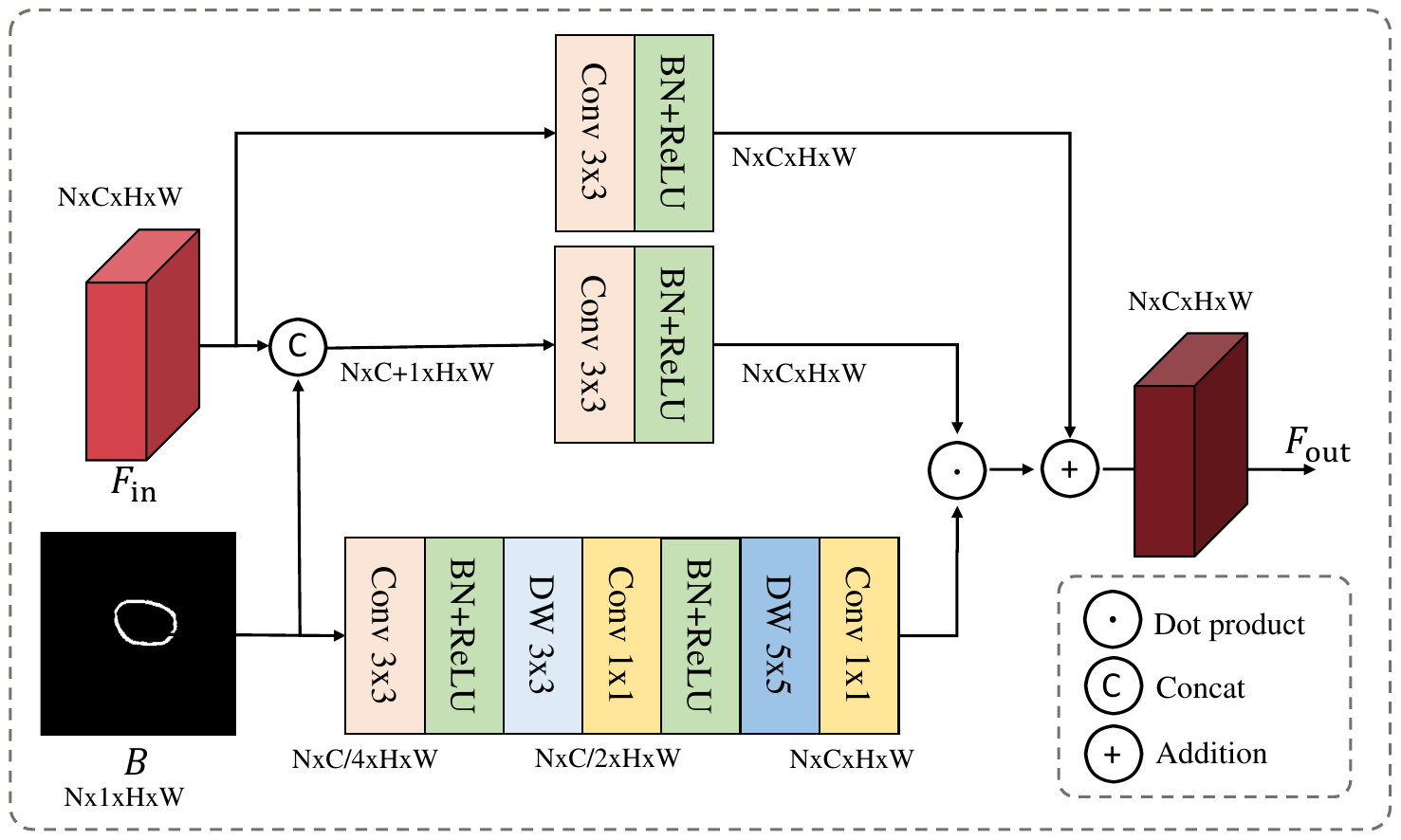} 
    \caption{The architecture of boundary-guided feature enhancement module. First, we will concatenate the input features and the boundary together, and pass the fused features into a $3\times3$ convolution network. Then, we operate multi-scale depth-wise convolutions for the boundary to expand the boundary information. Finally, we combine the expanded boundary and the enhanced features together through dot product operation. } 
	\label{fig:picBGFE}  
\end{figure}
\begin{equation}
F_{fused}=ReLU\left(BN\left(Conv3\times3\left(\left[F_{in};B\right]\right)\right)\right),
\label{eqa:eqfused}
\end{equation}
where ${Conv}{3\times3}$ denotes a $3\times 3$ convolution with  $C+1$ input channels and $C$ output channels. 

The boundary map $B$ is transformed into a spatial attention mask $Att \in \mathbb{R}^{H\times W\times C}$ through the multi-scale operations as follow:

\begin{equation}
B_{en}= CBR1\times1(DW3\times3(CBR3\times3 (B))),
\end{equation}
\begin{equation}
Att=Conv 1\times 1(DW5\times5(B_{en})),
\end{equation}
where $CBR3\times3$ and $CBR1\times1$ are defined as the sequential operation of a $3\times3$ or $1\times1$ convolution followed by batch normalization and a ReLU activation function. Likewise, we denote $DW3\times3$ and $DW5\times5$ as depthwise convolutions with kernel sizes of $3\times3$ and $5\times5$ respectively. Through the multi-scale convolution operations, the single-channel boundary is transformed into attention maps with $C$ dimensions. 



Finally, we get the enhanced features $F_{out}$ by element-wise multiplying the attention map $Att$ and the fused features $F_{fused}$, with a residual connection from original feature $F_{in}$, as follow:
\begin{equation}
F_{out}=Att\odot F_{fused} + CBR3\times3(F_{in}),
\end{equation}
where $\odot$ denoting element-wise multiplication, $F_{out}$ means the final enhanced features. The residual connection could preserve the original feature fidelity as well as augmenting boundary-critical information. 

By integrating boundary detection with adaptive feature enhancement, our BGFE module produces highly discriminative representations for ambiguous tumor margins. This mechanism substantially boosts segmentation precision in challenging ultrasound cases with low contrast and ambiguous boundaries.

\subsection{Scale-Aware Aggregation Module}

\begin{figure}[!t]
	\centering  
	\includegraphics[width=1\linewidth]{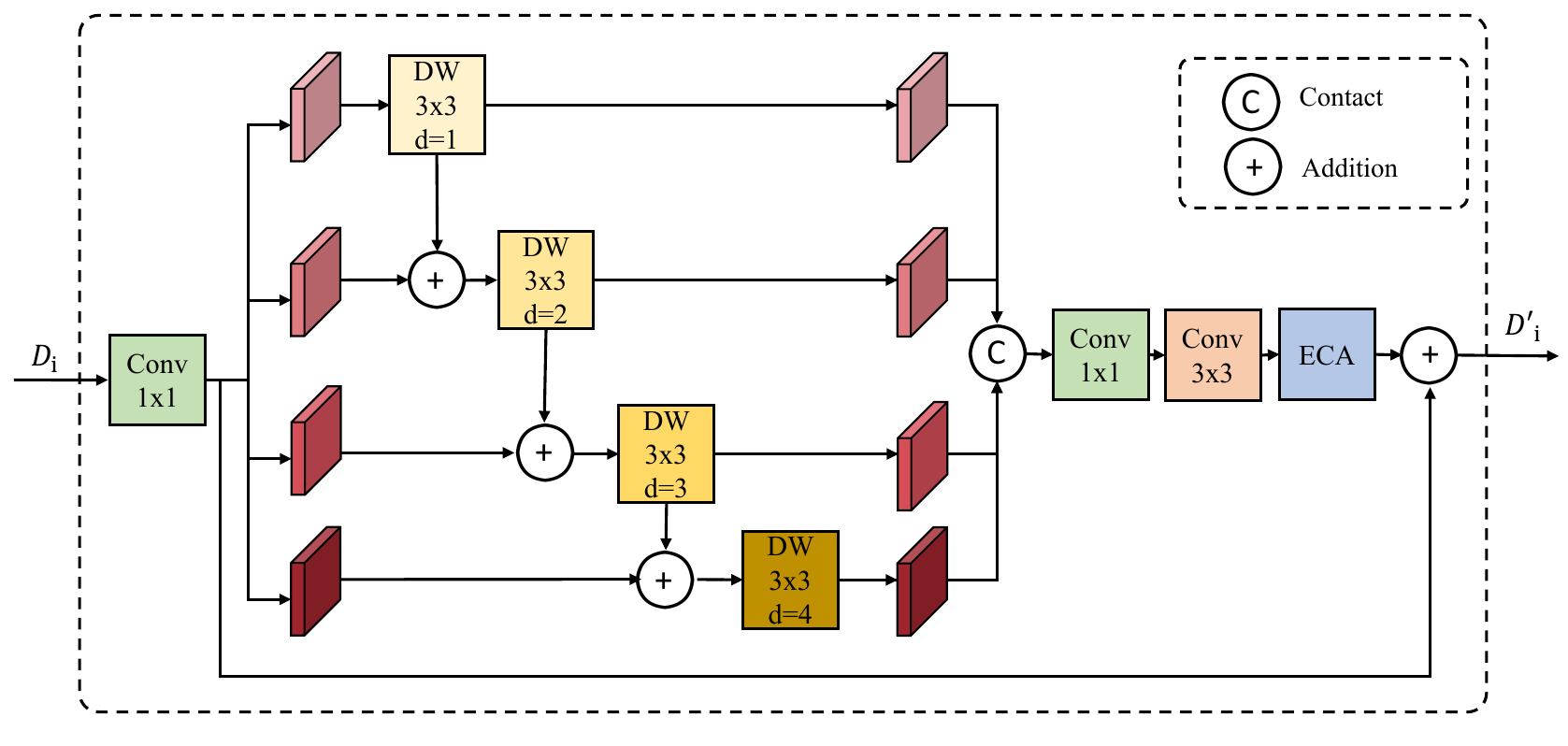} 
    \caption{The architecture of scale-aware aggregation module (SAAM). We split the input features into four groups along the channel dimension. Then, we put the four groups into depth-wise convolutions with different dilation rates. Then, we concatenate the multi-scale features and utilize a ECA module to adaptively adjust the representations.} 
	\label{fig:picSAAM}  
\end{figure}

The scale variability of lesions and low contrast of ultrasound image segmentation pose critical challenges for precise segmentation. To address these challenges, we propose the scale-aware aggregation module (SAAM), as shown in Fig.~\ref{fig:picSAAM}. We employ multi-scale depth-wise convolutions to enrich feature representations.

Here we take the $i$-th feature from decoder $D_{i}\in \mathbb{R}^{H\times W\times C}$ as an example. We first conduct channel dimension reduction via a $1\times1$ convolution, which is followed by batch normalization and a ReLU activation function, to obtain $D_{i,m}\in \mathbb{R}^{H\times W\times C'}$. Next, we evenly divide $D_{i,m}$ into four feature maps alone channel dimension, as follows:
\begin{equation}
D_{i,m}
\xrightarrow{\mathrm{split}} [D_{i,m}^1, D_{i,m}^2, D_{i,m}^3, D_{i,m}^4].
\end{equation}

Then, we perform cross-scale interaction learning, specifically integrating features from adjacent branches, and extracting multi-scale contextual features through a series of depthwise dilated convolutions. As shown in Fig.~\ref{fig:picSAAM}, these operations can be formulated as:
\begin{equation}
D_{i,m}^{1'}=DW(D_{i,m}^{1}),
\end{equation}
\begin{equation}
D_{i,m}^{2'}=DW(D_{i,m}^{1'}\oplus D_{i,m}^{2}), 
\end{equation}
\begin{equation}
D_{i,m}^{3'}=DW(D_{i,m}^{2'}\oplus D_{i,m}^{3}), 
\end{equation}
\begin{equation}
D_{i,m}^{4'}=DW(D_{i,m}^{3'}\oplus D_{i,m}^{4}), 
\end{equation}
where $DW(\cdot)$ indicates a $3\times3$ depthwise convolution with a dilation rate of $\{1,2,3,4\}$. $D_{i,m}^{1'}, D_{i,m}^{2'},D_{i,m}^{3'},D_{i,m}^{4'}$ are the feature maps with different scales. Then, we concatenate these multi-scale feature maps along the channel dimension, following with a $1\times1$ and $3\times3$ convolution to aggregate the multi-scale feature maps. In addition, we employ the Efficient Channel Attention (ECA)~\cite{wang2020eca} module to adaptively enhance the representations. Finally, we get the output with a residual connection from the input. 

Through the SAAM module, we can incorporate a semantic guidance strategy to enhance the discriminability of object boundaries. By establishing a correlation between local textures and global semantics, the module facilitates the interaction of information between neighboring receptive fields, effectively mitigating the semantic gap across different layers. This adaptive refinement process forces the model to focus on the intrinsic structure of the target, leading to more precise localization and complete segmentation results.

\subsection{Optimization}

Follow previous work~\cite{xue2021global}, we design a multi-task loss to simultaneously optimize the breast tumor segmentation and the boundary detection tasks. It combines the breast tumor segmentation loss and the boundary detection loss together. This multi-task loss design allows the segmentation model to learn both the segmentation information and boundary information of breast tumors.

\subsubsection{Segmentation Loss} 
In this paper, we employ the the Dice loss and Binary Cross Entropy (BCE) loss to optimize the segmentation loss. The segmentation loss consists of two terms, as follows:
\begin{equation}
L_{\mathrm{SEG}}=L_{\mathrm{Dice}}+\lambda_1\cdot L_{\mathrm{BCE}},
\label{eqa:eqa4}
\end{equation}
where $\lambda_1$ is the hyper-parameter to balance the effect of the Dice loss and BCE loss. Following~\cite{tang2023cmu,tang2024cmunext}, we set $\lambda_1=0.5$ for all experiments.

The Dice loss $L_{\mathrm{Dice}}$ between the prediction output and the ground truth is formulated as follows:
\begin{equation}
L_{\mathrm{Dice}}=1-\frac{2\sum_{i=1}^{H\cdot W}\hat{y}_{i}\cdot y_{i}}{\sum_{i=1}^{H \cdot W}\hat{y}^2_{i}+\sum_{i=1}^{H\cdot W}y^2_{i}},
\label{eqa:eqa5}
\end{equation}
where $\hat{y}_{i}$ is the predicted probability at the $i$-th pixel, and $y_{i}$ is the ground truth. Dice Loss could balance the effect of the foreground and the background.

The BCE loss $L_{\mathrm{BCE}}$ is calculated as follows:
\begin{equation}
L_{\mathrm{BCE}}=-\frac{1}{N}\sum_{i=1}^{H\cdot W}(y_{i}\log\hat{y}_{i})+(1-y_{i})\log(1-{\hat{y}}_{i}),
\label{eqa:eqa4-2}
\end{equation}
where $N=H\times W$ is the height and width of the input image. The BCE loss is able to measure the difference between the predicted probability and the ground truth label on a pixel-by-pixel basis so that the gradient propagates stably.

\subsubsection{Boundary Loss.}
Let the model outputs boundary prediction results as $\{{\hat{b}}_k\}^{4}_{k=1}$. Each predicted boundary map is up-sampled to the size of the input by bilinear interpolation. The boundary detection loss $L_{\mathrm{boundary}}$ is calculated as follows:
\begin{equation}
L_{\mathrm{boundary}}=-\frac{1}{N}\sum_{i=1}^{H\cdot W}b_{i}\log{\hat{b}}_{i}+(1-b_{i})\log(1-{\hat{b}}_{i}).
\label{eqa:eqa7}
\end{equation}

The total boundary losses are the mean of the boundary loss at each feature scale of decoders.
\begin{equation}
L_{\mathrm{Boundary}}=\frac{1}{K}\sum_{k=1}^{K}L_{\mathrm{boundary},k},
\label{eqa:eqa8}
\end{equation}
where $K=4$ indicates the number of decoder stages.

\subsubsection{Total Loss.}
The multi-task loss is calculated as follows:
\begin{equation}
L_{\mathrm{total}}=L_{\mathrm{SEG}}+\lambda_2\cdot L_{\mathrm{Boundary}}, 
\label{eqa:eqa9}
\end{equation}
where $\lambda_2$ is another hyper-parameter to balance the effect of the segmentation loss and boundary loss. Through conductive experiments, we find that $\lambda_2=0.7$ achieves better performance (see Section:~\ref{subsubsection:lossweight} for detailed results).

\section{Dataset and implement details}
\subsection{Dataset}
In this paper, we conduct experiments on four public datasets, BUSI~\cite{al2019deep}, Dataset B~\cite{yap2017automated}, TN3K~\cite{gong2021multi}, and BP~\cite{ultrasound-nerve-segmentation}. The \textbf{BUSI dataset}~\cite{al2019deep} contains 780 ultrasound images of 600 women between the ages of 25 and 75 years. It includes 133 normal images, 437 benign tumors, and 210 malignant tumors. And the average image resolution is $500\times500$ pixels. Following recent studies~\cite{tang2023cmu, qin2025mbe}, we only utilize benign and malignant cases from this dataset. The training and validation sets are divided in $8:2$. The \textbf{Dataset B}~\cite{yap2017automated} consists of 163 samples from different patients, including 53 malignant and 110 benign samples. It is collected at the UDIAT Diagnostic Center of the Parc Tauli Corporation, Sabadell, Spain, using a Siemens ACUSON Sequoia C512 system, and annotated by experienced radiologists. The training and validation sets are divided in $8:2$. The \textbf{TN3k}~\cite{gong2021multi} consists of 3493 ultrasound images of thyroid nodules from 2421 patients. The dataset contributors have already partitioned the dataset, including 2879 images and 614 images for the training and validation. The \textbf{BP}~\cite{ultrasound-nerve-segmentation} dataset is from the Kaggle Ultrasound Nerve Segmentation Challenge. This dataset contains 2323 images with nerves and 3312 normal images. The resolution for images from BP dataset is $580\times240$. Following ~\cite{qin2025mbe}, we selected the 2323 images with nerves to do the experiments, dividing the 2323 images into training, validation, and test with a ratio of $7:2:1$.
\subsection{Experimental parameterization}
We employ SGD optimizer with the initial learning rate is 0.001, the weight decay is 0.0001, and the momentum is 0.9. We adopt the ``Poly'' decay strategy to adjust the learning rate. For BUSI, Dataset B, TN3K and BP, the batch size is 8, 8, 16 and 16, the training epoch number is 300, 300, 40 and 40, and the image were resized to $256\times256$, $256\times256$, $224\times224$ and $224\times224$, respectively. Our experimental device is a PC with single NVIDIA RTX 4090 GPU. The development environment is Ubuntu 20.04, python 3.8 and PyTorch 1.13.



\subsection{Evaluation Metric}

We adopt five commonly used metrics to quantitatively evaluate the performance of different segmentation models: Dice, Intersection over Union (IoU), Hausdorff Distance 95\% (HD95), Recall, and Accuracy.
\begin{equation}
\text{Dice} = \frac{\text{2TP}}{\text{2TP} + \text{FP} + \text{FN}},
\label{eqa:eqaDice}
\end{equation}
\begin{equation}
\text{IoU} = \frac{\text{TP}}{ \text{TP} + \text{FN} + \text{FP}},
\label{eqa:eqaIOU}
\end{equation}
\begin{equation}
\text{HD95} = \max_{k95\%}\left[ d(X,Y), d(Y,X) \right]
\end{equation}
\begin{equation}
\text{Recall} = \frac{\text{TP}}{\text{TP} + \text{FN}},
\label{eqa:eqaRecall}
\end{equation}

\begin{equation}
\text{Accuracy} = \frac{\text{TP} + \text{TN}}{\text{TP} + \text{TN} + \text{FP} + \text{FN}},
\label{eqa:eqaAcc}
\end{equation}
where TP is defined as the positive output for the corresponding positive ground truth (GT). FP is defined as the positive  output for the corresponding negative GT. TN means the negative output for the corresponding negative GT. FN means the  negative output for the corresponding positive GT. For the segmentation results, higher values of metrics such as the Dice, IoU, Recall, and Accuracy indicate better performance, whereas a lower value of HD95 is desirable.

\section{Experimental results}
\subsection{Comparisons with State-of-the-Art Methods}
In this section, we compare our method against several state-of-the-art segmentation methods to validate the effectiveness of our proposed method. First, we compare our method with some famous traditional medical image segmentation methods, including U-Net~\cite{ronneberger2015u}, U-Net++~\cite{zhou2019unet++}, Attention U-Net~\cite{oktay2018attention}, FCN-8s~\cite{long2015fully}, DeeplabV3+~\cite{chen2018encoder}, and TransUNet~\cite{chen2021transunet}. Then, we compare our method with some ultrasound segmentation methods, such as AAU-Net~\cite{chen2022aau}, ESKNet~\cite{chen2024esknet}, CMUNet~\cite{tang2023cmu}, CMUNeXt~\cite{tang2024cmunext}, GFA-Net~\cite{GFA-Net}, PConv-UNet~\cite{wang2025pconv}, Syn-Net (JBHI2025)~\cite{zhao2024syn}, and LGFFM (TMI2025)~\cite{luo2025lgffm}. In addition, we also compare our method with some famous boundary-assisted segmentation methods, GGNet~\cite{xue2021global}, BGNet~\cite{sun2022bgnet}, BGRA-GSA~\cite{BGRA-GSA23}, BFNet~\cite{yue2025boundary}, MBE-UNet~\cite{qin2025mbe}, and CTO~\cite{lin2025rethinking}. Table~\ref{tab:tabBUSI}-\ref{tab:tabBP} present the quantitative comparison results with the baseline models, and Fig.~\ref{fig:picVisulizationBUSI}-\ref{fig:picVisulizationBP} present the qualitative comparison results.


\subsubsection{\textbf{Comparison Results on BUSI Dataset}}
Table.~\ref{tab:tabBUSI} presents the compared results with some state-of-the-art methods on BUSI dataset. As shown in Table~\ref{tab:tabBUSI}, ultrasound-specific segmentation methods generally outperform general medical image segmentation approaches. Furthermore, boundary-assisted methods tend to achieve better results than ultrasound-specific segmentation methods in several cases. It can be observed that our proposed PBE-UNet method outperforms state-of-the-art methods across all the five evaluation metrics. Specifically, our model attains 85.34\% in Dice, 77.52\% in IoU, 11.66 mm in HD95, 87.95\% in Recall and 97.16\% in Accuracy. Compared to U-Net, our method shows an improvement of 9.77\% in Dice, 10.14\% in IoU, 10.74\% in Recall, and 1.06\% in Accuracy, while reducing HD95 by 11.95 mm. Compared to the recent method CTO, our model still surpasses it by 1.55\% in Dice, 1.56\% in IoU, 2.18\% in Recall, and 0.43\% in Accuracy, while also achieving a reduction of 1.85 mm in HD95. These results demonstrate the effectiveness of our proposed method.

\begin{figure*}[!t]
	\centering  
	\includegraphics[width=\linewidth]{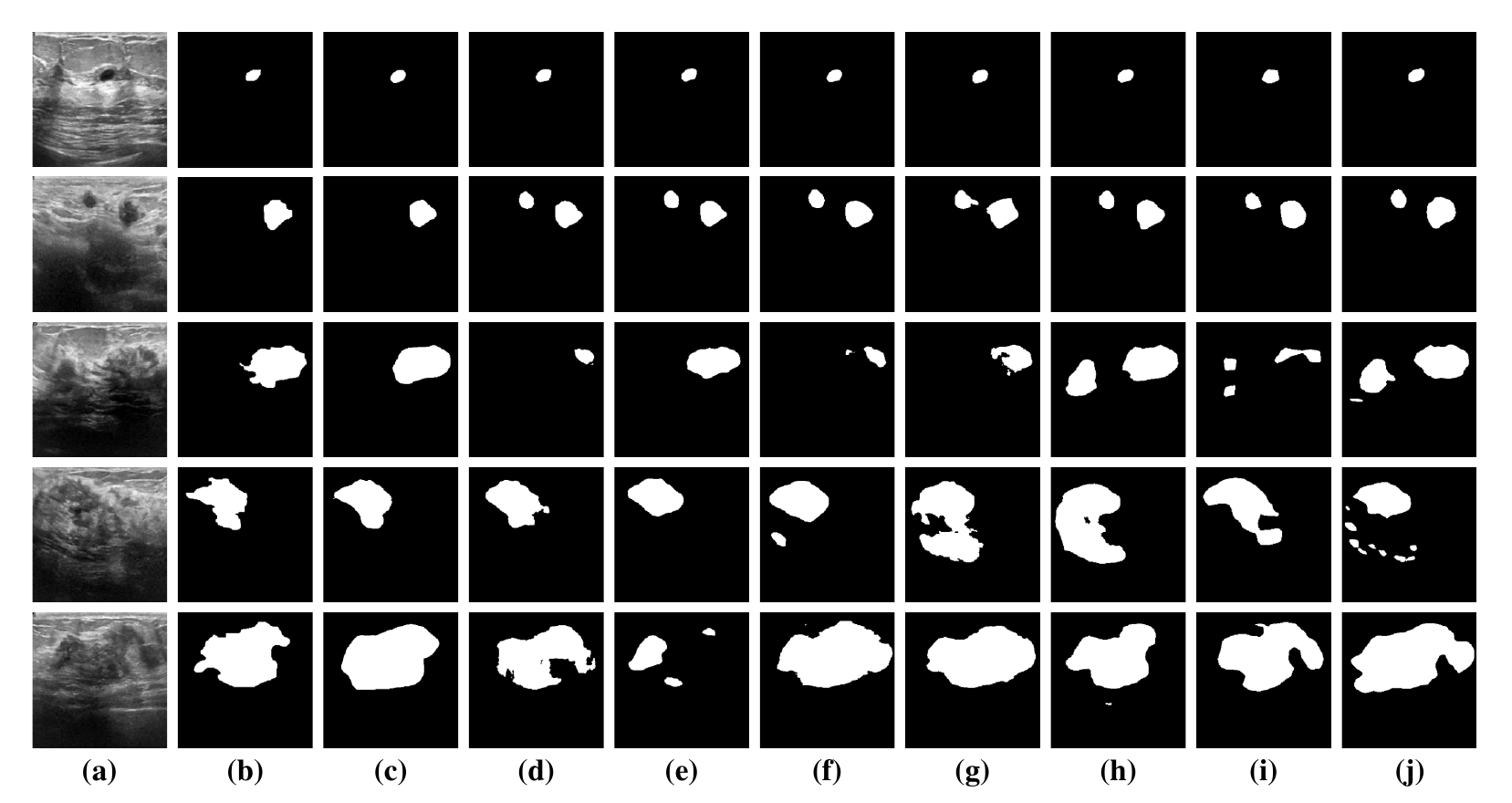} 
	\caption{Visual comparison of segmentation results of the proposed model with seven state-of-the-art methods on the BUSI dataset. (a) original image, (b) GT, (c) ours, (d) U-Net, (e) TransUNet, (f) CMUNet, (g) CMUNeXt, (h) GGNet, (i) BGNet, (j) BFNet.} 
        \label{fig:picVisulizationBUSI}
\end{figure*}

In Fig.~\ref{fig:picVisulizationBUSI}, we visually compare the segmentation results of the PBE-UNet with some SOTA segmentation models on the BUSI dataset. It is evident that our method effectively delineates the lesion regions with more precise localization, successfully distinguishing pathological tissue from surrounding areas. In contrast, most comparative models produce segmentation with blurred or irregular contours, struggling to capture low-contrast regions and fine morphological details.

\begin{table}[!t]
\caption{Results on the BUSI dataset.}
\centering
\resizebox{1\linewidth}{!}{
\begin{tabular}{lccccc}
\toprule
Model&Dice$\%\color{red}{\uparrow}$&IoU$\%\color{red}{\uparrow}$&HD95$\color{red}{\downarrow}$&Recall$\%\color{red}{\uparrow}$&Acc$\%\color{red}{\uparrow}$ \\
\midrule
\multicolumn{6}{l}{\textcolor{blue!90}{\emph{Traditional Segmentation Methods}}}\\
U-Net (MICCAI2015)~\cite{ronneberger2015u} &75.57 &67.38 &23.61 &77.21 &96.10\\
FCN-8s (CVPR2015)~\cite{long2015fully} &79.51 &70.58 &20.50 &81.24 &96.01\\
DeeplabV3+ (ECCV2018)~\cite{chen2018encoder} &81.31 &73.19 &15.73 &84.30 &96.25\\
AttUNet (MIDL2018)~\cite{oktay2018attention} &79.59 &71.10 &21.44  &81.77 &95.85\\
U-Net++ (TMI2020)~\cite{zhou2019unet++} &79.11 &70.32 &24.04 &83.06 &96.87\\
TransUNet (ARXIV2021)~\cite{chen2021transunet} &82.94 &74.79 &16.37 &84.80 &96.75\\

\midrule
\textcolor{blue!90}{\emph{Ultrasound Segmentation Methods}}\\
AAU-net (TMI2022)~\cite{chen2022aau} &77.51 &68.82 &22.47 &81.10 &96.77 \\
CMUNet (ISBI2023)~\cite{tang2023cmu} &82.62 &74.64 &14.79 &82.80 &97.11\\
CMUNeXt (ISBI2024)~\cite{tang2024cmunext}&83.05 &74.91 &14.34 &83.80 &96.95 \\
GFA-Net (TIM2025)~\cite{GFA-Net}&79.99 &70.73 &-- &-- &-- \\
PConv-UNet (DISPLAY2025)~\cite{wang2025pconv}&84.48 &73.76 &-- &82.70 &-- \\
Syn-Net (JBHI2025)~\cite{zhao2024syn}&82.74 &74.59 &-- &85.50 &-- \\
LGFFM (TMI2025)~\cite{luo2025lgffm} &79.99 &71.80 &-- &-- &-- \\

\midrule
\textcolor{blue!90}{\emph{Boundary-Assisted Methods}}\\
GGNet (MIA2021)~\cite{xue2021global} &82.63 &74.23 &14.50 &87.46 &96.63 \\
BGNet (IJCAI2022)~\cite{sun2022bgnet} &83.05 &74.98 &14.22 &84.20 &96.98 \\
BGRA-GSA (JBHI2023)~\cite{BGRA-GSA23} &81.43 &68.75 &-- &-- &96.34\\ 
BFNet (TCSVT2025)~\cite{yue2025boundary} &81.45 &73.32 &16.82 &82.30 &96.66 \\
MBE-UNet (JBHI2025)~\cite{qin2025mbe} &84.14 &75.02 &14.12 &85.45 &96.16 \\
CTO (MIA2025)~\cite{lin2025rethinking} &83.79 &75.96 &13.51 &85.77 &96.73\\
\rowcolor{blue!10}\textbf{PBE-UNet (ours)} &\textbf{85.34} &\textbf{77.52} &\textbf{11.66} &\textbf{87.95} &\textbf{97.16}\\
\bottomrule
\end{tabular}}
\label{tab:tabBUSI}
\end{table}
\begin{table}[!t]
\caption{Results on the Dataset B dataset. }
\centering
\resizebox{1\linewidth}{!}{
\begin{tabular}{lccccc}
\toprule
Model&Dice$\%\color{red}{\uparrow}$&IoU$\%\color{red}{\uparrow}$&HD95$\color{red}{\downarrow}$&Recall$\%\color{red}{\uparrow}$&Acc$\%\color{red}{\uparrow}$ \\
\midrule
\multicolumn{6}{l}{\textcolor{blue!90}{\emph{Traditional Segmentation Methods}}}\\
U-Net (MICCAI2015)~\cite{ronneberger2015u} &80.29 &72.90 &14.69 &78.99 &98.44\\
AttUNet (MIDL2018)~\cite{oktay2018attention} &79.30 &72.93 &14.44  &76.15 &97.85\\
U-Net++ (TMI2020)~\cite{zhou2019unet++} &80.73 &71.04 &14.80 &78.26 &97.97 \\
TransUNet (ARXIV2021)~\cite{chen2021transunet} &83.77 &75.28 &12.56 &84.11 &97.92\\
\midrule
\textcolor{blue!90}{\emph{Ultrasound Segmentation Methods}}\\
AAU-net (TMI2022)~\cite{chen2022aau} &81.29 &72.74 &13.70 &82.05 &97.68 \\
CMUNet (ISBI2023)~\cite{tang2023cmu} &84.20 &75.78 &8.19 &82.71 &98.45\\
ESKNet (ESWA2024)~\cite{chen2024esknet}&82.10 &73.41 &14.81 &85.98 &98.66\\
CMUNeXt (ISBI2024)~\cite{tang2024cmunext} &82.85 &74.14 &9.68 &82.36 &98.25\\
GFA-Net (TIM2025)~\cite{GFA-Net}&85.60 &76.70 &-- &-- &-- \\
PConv-UNet (DISPLAY2025)~\cite{wang2025pconv}&87.61 &78.20 &-- &\textbf{91.66} &-- \\
Syn-Net (JBHI2025)~\cite{zhao2024syn} &86.92 &78.80 &6.89 &90.98 &98.47 \\
LGFFM (TMI2025)~\cite{luo2025lgffm} &83.80 &75.40 &-- &-- &-- \\
\midrule
\textcolor{blue!90}{\emph{Boundary-Assisted Methods}}\\
GGNet (MIA2021)~\cite{xue2021global} &84.80 &76.04 &9.10 &91.16 &98.44 \\
BGRA-Net (BIBM2021)~\cite{BGRA-Net21} &86.63 &76.69 &-- &-- &98.75 \\
BGNet (IJCAI2022)~\cite{sun2022bgnet} &87.63 &79.65 &6.66 &90.32 &98.73 \\
BGRA-GSA (JBHI2023)~\cite{BGRA-GSA23} &88.01 &78.80 &-- &-- &\textbf{98.80} \\
BFNet (TCSVT2025)~\cite{yue2025boundary} &82.79 &73.89 &10.35 &82.57 &98.01 \\
MBE-UNet (JBHI2025)~\cite{qin2025mbe} &87.30 &79.06 &6.38 &88.84 &98.49 \\
\rowcolor{blue!10}\textbf{PBE-UNet (ours)} &\textbf{88.78} &\textbf{81.46} &\textbf{5.29} &91.56 &98.79\\
\bottomrule
\label{tab:tabDatasetB}
\end{tabular}}
\end{table}

\subsubsection{\textbf{Comparison Results on Dataset B}}
Table.~\ref{tab:tabDatasetB} presents the results on Dataset B. We can find the similar results that ultrasound-specific segmentation methods generally outperform general medical image segmentation approaches. Results in Table.~\ref{tab:tabDatasetB} shows that our model attains leading performance on the majority of key metrics, Dice, IoU and HD95. In specific, our proposed method achieves the best scores of 88.78\% in Dice, 81.46\% in IoU, and 5.29 mm in HD95, while also delivering highly competitive results of 91.56\% in Recall and 98.79\% in Accuracy. Compared to the classic U-Net, our method shows substantial improvement across all metrics, for instance, increasing the Dice by 8.49\%, the IoU by 8.56\% and remarkably reducing HD95 by 9.40 mm.

\begin{figure*}[!t]
	\centering  
	\includegraphics[width=\linewidth]{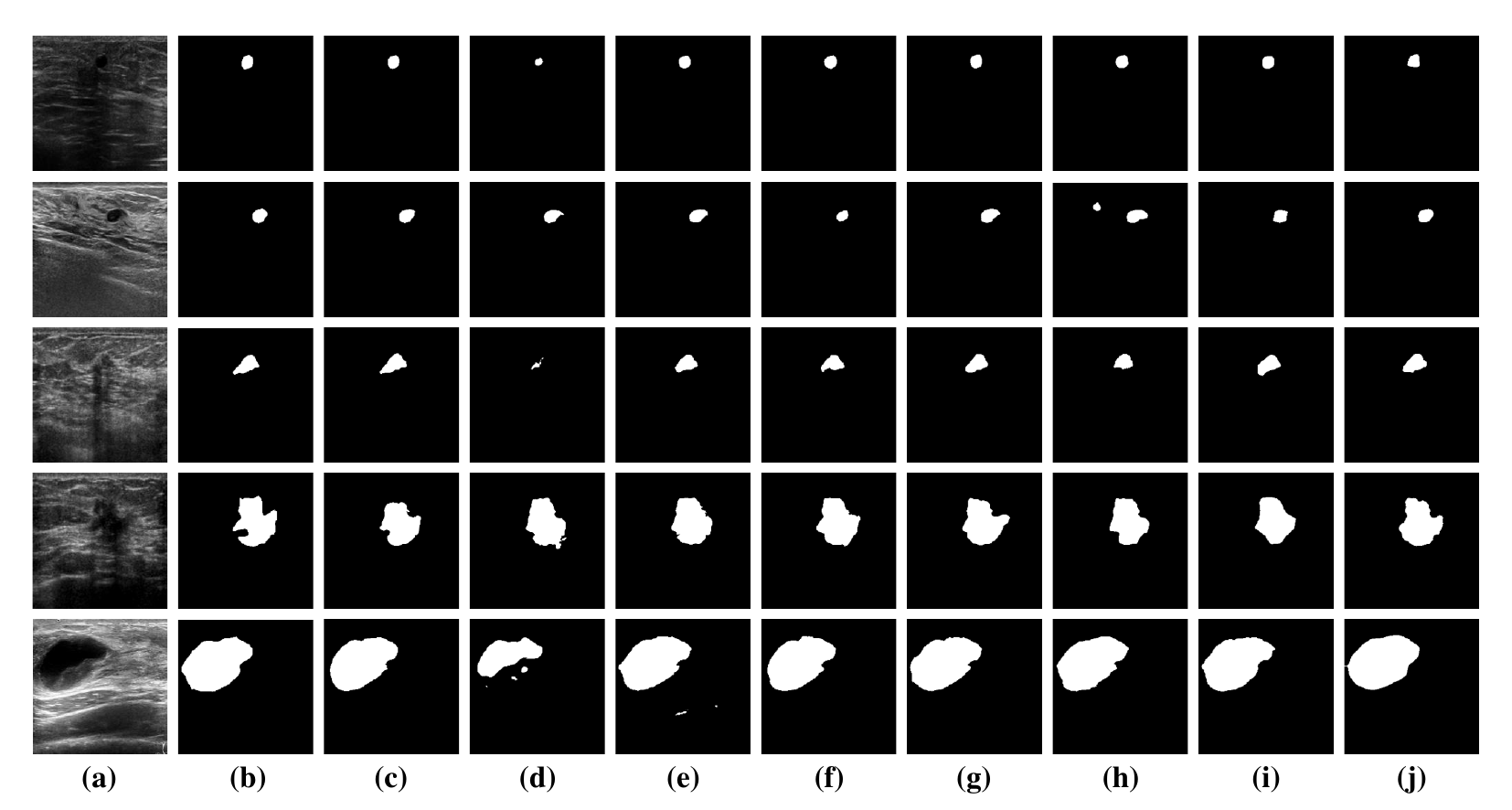} 
	\caption{Visual comparison of segmentation results of the proposed model with seven state-of-the-art methods on the Dataset B dataset. (a) original image, (b) GT, (c) ours, (d) U-Net, (e) TransUNet, (f) CMUNet, (g) CMUNeXt, (h) GGNet, (i) BGNet, (j) BFNet.} 
	\label{fig:picVisulizationDatasetB}  
\end{figure*}

We visualize the tumor segmentation results on dataset B in Fig.~\ref{fig:picVisulizationDatasetB}. It can be observed that our method accurately delineates the complete and continuous contour of targets, maintaining smooth boundaries even in challenging regions with blurry edges or low contrast.

\begin{figure*}[!t]
	\centering  
	\includegraphics[width=\linewidth]{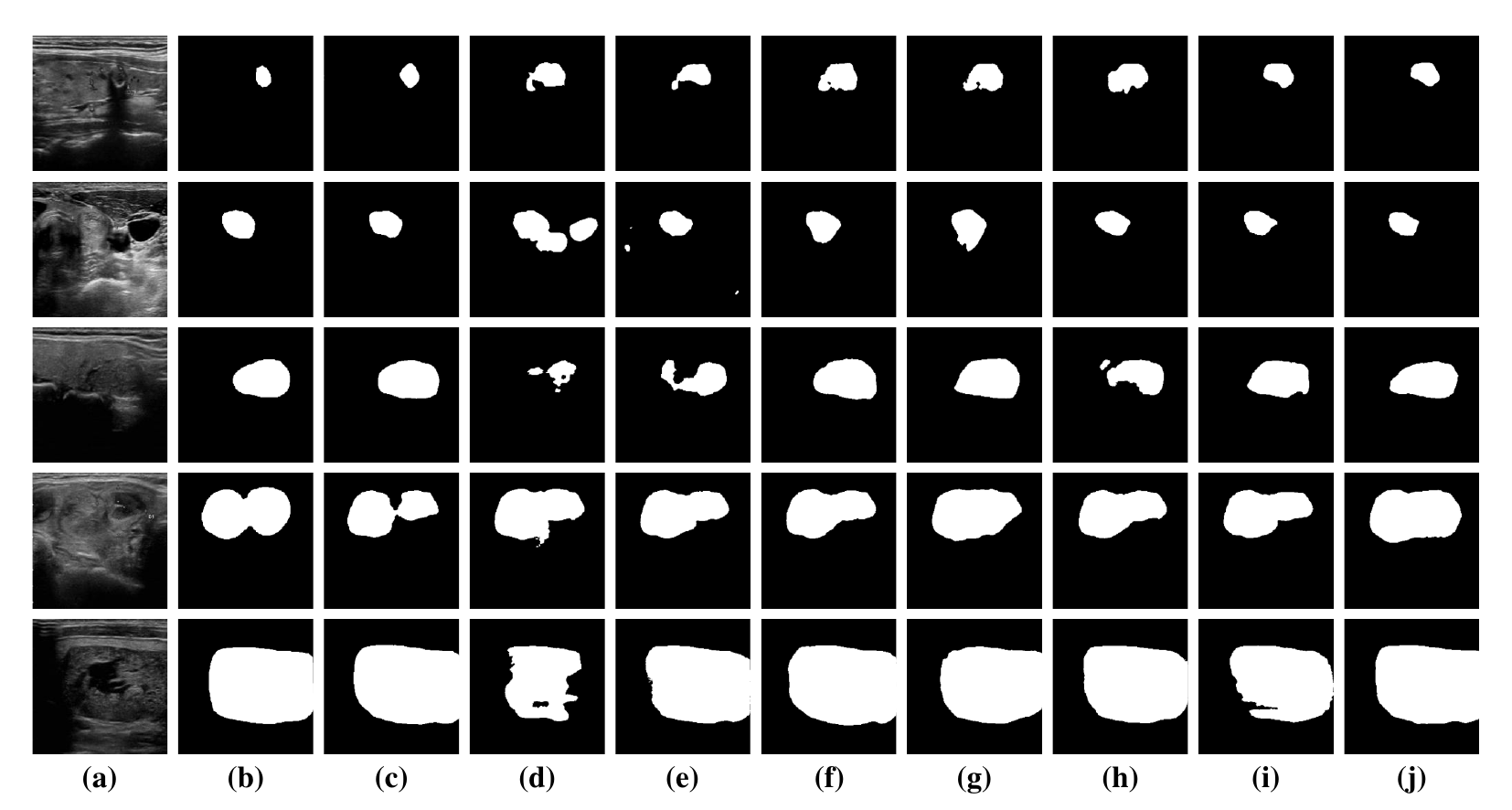} 
	\caption{Visual comparison of segmentation results of the proposed model with seven state-of-the-art methods on the TN3K dataset. (a) original image, (b) GT, (c) ours, (d) U-Net, (e) TransUNet, (f) CMUNet, (g) CMUNeXt, (h) GGNet, (i) BGNet, (j) BFNet.} 
	\label{fig:picVisulizationTN3K}  
\end{figure*}

\begin{table}[!t]
\caption{Results on the TN3K dataset.}
\centering
\resizebox{1\linewidth}{!}{
\begin{tabular}{lccccc}
\toprule
Model&Dice$\%\color{red}{\uparrow}$&IoU$\%\color{red}{\uparrow}$&HD95$\color{red}{\downarrow}$&Recall$\%\color{red}{\uparrow}$&Acc$\%\color{red}{\uparrow}$ \\
\midrule
\multicolumn{6}{l}{\textcolor{blue!90}{\emph{Traditional Segmentation Methods}}}\\
U-Net (MICCAI2015)~\cite{ronneberger2015u} &76.60 &65.91 &19.38 &82.71 &96.79\\
FCN-8s (CVPR2015)~\cite{long2015fully} &78.07 &67.69 &16.24 &82.63 &96.91\\
DeeplabV3+ (ECCV2018)~\cite{chen2018encoder} &79.66 &69.61 &14.52 &84.60 &96.90\\
AttUNet (MIDL2018)~\cite{oktay2018attention} &79.01 &69.29 &16.13 &83.61 &96.85 \\
U-Net++ (TMI2020)~\cite{zhou2019unet++} &78.85 &68.99 &16.55 &82.47 &96.87 \\
TransUNet (ARXIV2021)~\cite{chen2021transunet} &78.97 &68.68 &15.03 &83.05 &96.79\\
\midrule
\textcolor{blue!90}{\emph{Ultrasound Segmentation Methods}}\\
AAU-net (TMI2022)~\cite{chen2022aau} &78.23 &68.74 &16.77 &82.82 &96.81 \\
CMUNet (ISBI2023)~\cite{tang2023cmu} &82.68 &74.10 &12.34 &85.39 &97.06\\
CMUNeXt (ISBI2024)~\cite{tang2024cmunext}&82.12 &73.21 &12.42 &87.07 &97.01 \\
LGFFM (TMI2025)~\cite{luo2025lgffm} &81.76 &71.23 &-- &-- &-- \\
\midrule
\textcolor{blue!90}{\emph{Boundary-Assisted Methods}}\\
GGNet (MIA2021)~\cite{xue2021global} &81.72 &72.16 &11.95 &84.62 &96.62 \\
BGNet (IJCAI2022)~\cite{sun2022bgnet} &80.74 &71.59 &14.13 &86.93 &96.77 \\
BPAT-UNet (CMPB2023)~\cite{bi2023bpat} &80.14 &70.70 &16.60 &85.55 &96.89 \\
BFNet (TCSVT2025)~\cite{yue2025boundary} &83.10 &74.38 &11.44 &86.62 &97.09 \\
MBE-UNet (JBHI2025)~\cite{qin2025mbe} &81.16 &71.72 &12.61 &85.24 &97.15 \\
CTO (MIA2025)~\cite{lin2025rethinking} &82.31 &73.73 &12.52 &86.48 &96.86\\
\rowcolor{blue!10}\textbf{PBE-UNet (ours)} &\textbf{83.43} &\textbf{75.02} &\textbf{10.95} &\textbf{88.64} &\textbf{97.19}\\
\bottomrule
\label{tab:tabTN3K}
\end{tabular}}
\end{table}
\begin{table}[!t]
\caption{Results on the BP dataset.}
\centering
\resizebox{1\linewidth}{!}{
\begin{tabular}{lccccc}
\toprule
Model&Dice$\%\color{red}{\uparrow}$&IoU$\%\color{red}{\uparrow}$&HD95$\color{red}{\downarrow}$&Recall$\%\color{red}{\uparrow}$&Acc$\%\color{red}{\uparrow}$ \\
\midrule
\multicolumn{6}{l}{\textcolor{blue!90}{\emph{Traditional Segmentation Methods}}}\\
U-Net (MICCAI2015)~\cite{ronneberger2015u} &78.13 &66.91 &10.44 &80.76 &98.71\\
FCN-8s (CVPR2015)~\cite{long2015fully} &79.05 &67.74 &8.62 &81.01 &98.79\\
DeeplabV3+ (ECCV2018)~\cite{chen2018encoder} &79.74 &67.97 &7.94 &82.09 &98.79\\
AttUNet (MIDL2018)~\cite{oktay2018attention} &78.76 &67.52 &9.84 &80.68 &98.80\\
U-Net++ (TMI2020)~\cite{zhou2019unet++} &79.65 &68.50 &8.94 &80.80 &98.82\\
TransUNet (ARXIV2021)~\cite{chen2021transunet} &79.25 &67.79 &9.64 &81.72 &98.79\\
\midrule
\textcolor{blue!90}{\emph{Ultrasound Segmentation Methods}}\\
AAU-net (TMI2022)~\cite{chen2022aau} &79.00 &67.23 &9.36 &81.91 &98.76 \\
CMUNet (ISBI2023)~\cite{tang2023cmu} &79.88 &68.45 &8.74 &80.73 &98.85\\
CMUNeXt (ISBI2024)~\cite{tang2024cmunext}&79.77 &68.46 &8.79 &81.08 &98.81 \\
\midrule
\textcolor{blue!90}{\emph{Boundary-Assisted Methods}}\\
GGNet (MIA2021)~\cite{xue2021global} &78.33 &66.22 &8.27 &81.76 &98.64 \\
BGNet (IJCAI2022)~\cite{sun2022bgnet} &79.61 &68.19 &7.92 &81.60 &98.79 \\
BFNet (TCSVT2025)~\cite{yue2025boundary} &79.15 &68.66 &7.73 &82.68 &98.84 \\
MBE-UNet (JBHI2025)~\cite{qin2025mbe} &79.93 &68.85 &8.08 &82.47 &98.86 \\
CTO (MIA2025)~\cite{lin2025rethinking} &80.03 &68.97 &7.64 &81.79 &98.88\\
\rowcolor{blue!10}\textbf{PBE-UNet (ours)} &\textbf{81.02} &\textbf{69.96} &\textbf{6.27} &\textbf{83.15} &\textbf{98.89}\\
\bottomrule
\label{tab:tabBP}
\end{tabular}}
\end{table}
\begin{figure*}[!t]
	\centering  
	\includegraphics[width=\linewidth]{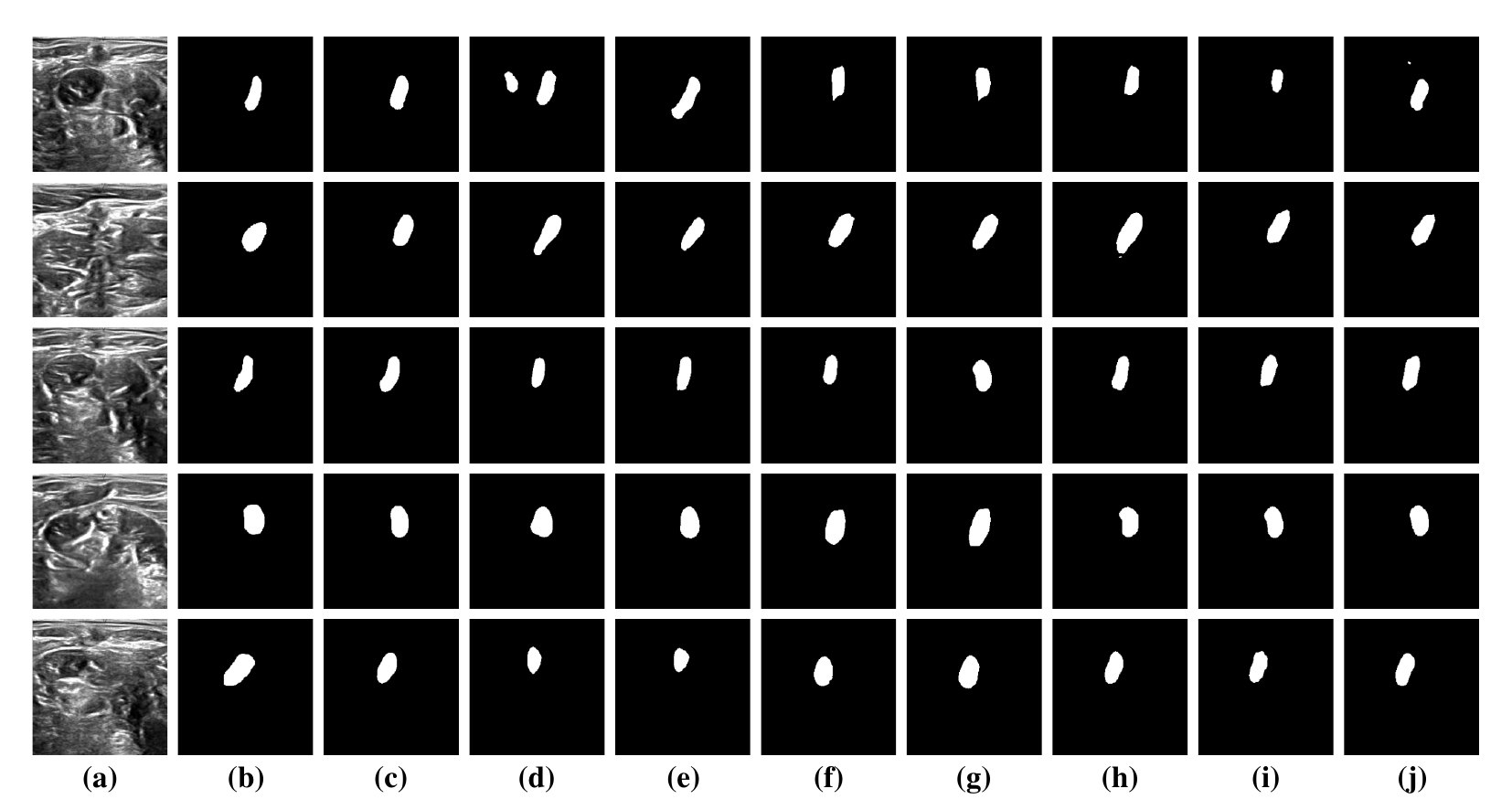} 
	\caption{Visual comparison of segmentation results of the proposed model with seven state-of-the-art methods on the BP dataset. (a) original image, (b) GT, (c) ours, (d) U-Net, (e) TransUNet, (f) CMUNet, (g) CMUNeXt, (h) GGNet, (i) BGNet, (j) BFNet.} 
	\label{fig:picVisulizationBP}  
\end{figure*}

\subsubsection{\textbf{Comparison Results on TN3K Dataset}}
Table.~\ref{tab:tabTN3K} presents the quantitative results on the TN3K dataset. Our proposed method still achieves the best scores across all five metrics on TN3K dataset. It attains Dice of 83.43\%, IoU of 75.02\%, HD95 of 10.95 mm, Recall of 88.64\%, and Accuracy of 97.19\%. Compared to the second best method BFNet, our model obtains improvements with respective leads of 0.33\% in Dice and 0.64\% in IoU, while also reducing HD95 by 0.49 mm. These results demonstrate the effectiveness of our proposed method.

Fig.~\ref{fig:picVisulizationTN3K} presents a qualitative comparison between our network and other state-of-the-art methods on the TN3K dataset. It is evident that our proposed method produces contours that most closely match the ground truth, with sharp boundaries and well-preserved nodule morphology. Conversely, most comparative networks struggle to refine the nodule boundaries, often resulting in segmentation with blurred or irregular edges.

\subsubsection{\textbf{Comparison Results on BP Dataset}}
Table.~\ref{tab:tabBP} presents the results of state-of-the-art methods on the BP dataset. It can be observed that our proposed PBE-UNet model achieves superior segmentation performance than other methods. It obtains the best results across all five evaluation metrics. Specifically, our model attains 69.96\% in IoU, 81.02\% in Dice, 6.27 mm in HD95, 83.15\% in Recall, and 98.89\% in Accuracy. Compared to the U-Net, our method shows improvements of 3.05\% in IoU, 2.89\% in Dice, 2.39\% in Recall, and 0.18\% in Accuracy, while reducing HD95 by 4.17 mm. When compared to the second-best performing model, CTO, our model still surpasses COT method by 0.99\% in IoU, 0.99\% in Dice, 1.36\% in Recall, and 0.01\% in Accuracy, while further lowering HD95 by 1.37 mm. These results collectively indicate that our model holds a significant advantage in enhancing both segmentation precision and boundary accuracy.

Fig.~\ref{fig:picVisulizationBP} illustrates the qualitative segmentation of different SOTA methods on the BP dataset. It can be observed that our PBE-UNet model yields segmentation masks that are visually closest to the ground truth annotations. Our approach could accurately delineate the complete contour of the targets, maintain boundary continuity and clarity, even in challenging areas with blurry edges or low contrast to the surrounding background tissue.



\begin{table}[!t]
\caption{ABLATION EXPERIMENT OF THE PROPOSED MODULES ON THE BUSI Dataset. BD means the boundary detection module, BGFE means the boundary-guided feature enhancement module, and SAAM means the adaptive aggregation module. }
\centering
\resizebox{\linewidth}{!}{
\begin{tabular}{cccccccc}
\toprule
BD & BGFE & SAAM &Dice$\%\color{red}{\uparrow}$&IoU$\%\color{red}{\uparrow}$&HD95$\color{red}{\downarrow}$&Recall$\%\color{red}{\uparrow}$&Acc$\%\color{red}{\uparrow}$ \\
\midrule
$\times$ &$\times$  &$\times$  & 83.05 &74.91 &14.34 &83.80 &96.95\\
\checkmark &$\times$  &$\times$ &83.61 &75.25 &14.25 &84.55 &\textbf{97.20} \\
\checkmark &\checkmark &$\times$ &84.95 &76.71 &12.91 &86.79 &96.85\\
$\times$ &$\times$ &\checkmark &83.35 &75.35 &14.02 &84.24 &97.07 \\
\checkmark &\checkmark &\checkmark &\textbf{85.34} &\textbf{77.52} &\textbf{11.66} &\textbf{87.95} &97.16\\
\bottomrule
\label{tab:tabAblation}
\end{tabular}}
\end{table}
\begin{figure}[!t]
    \centering
    \includegraphics[width=\linewidth]{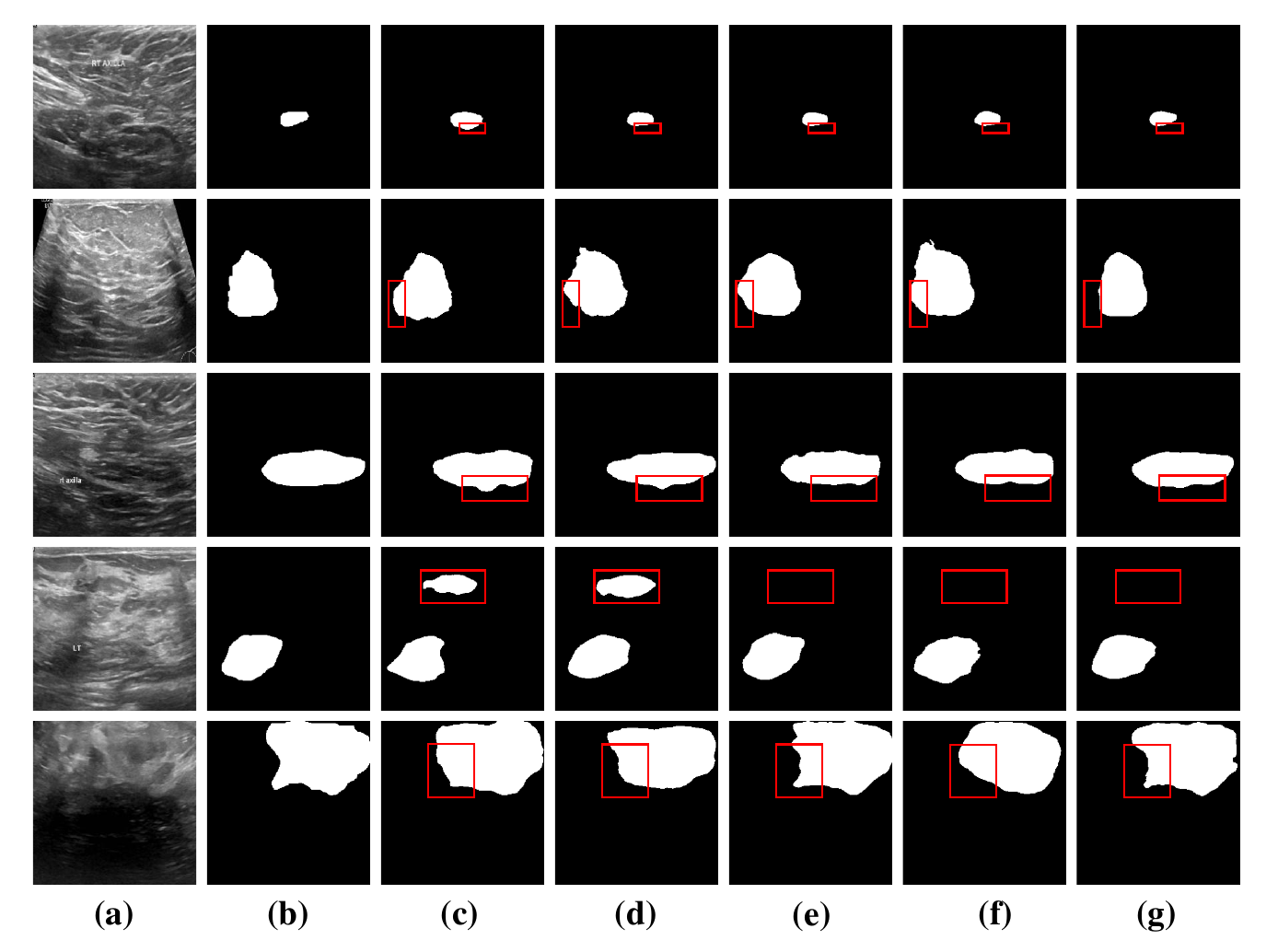}
    \caption{The visual segmentation results with different modules on the BUSI dataset. (a) original image, (b) GT, (c) baseline model, (d) baseline + BD, (e) baseline + BD + BGFE, (f) baseline + SAAM, (g) baseline + BD + BGFE + SAAM.}
    \label{fig:FigAblation}
\end{figure}

\subsection{Ablation Studies}
\label{sec:ablation}
\subsubsection{Effectiveness of each component }

To validate the effectiveness of the proposed boundary detection (BD), boundary-guided feature enhancement (BGFE) module , and scale-aware aggregation module (SAAM), we conducted ablation studies and reported the results in Tab.~\ref{tab:tabAblation}. From the results in Tab.~\ref{tab:tabAblation}, we can find that the baseline model can achieve 83.05\% in Dice, 74.91\% in IoU, and 14.34 mm in HD95. Then, the introduction of the BD module yielded slight improvements as the Dice increases by 0.56\% and the IoU by 0.34\%, while the HD95 decreases by 0.09 mm. This result validates the effectiveness of explicit boundary detection. Furthermore, the BGFE module significantly boosts the performance by increasing the Dice and IoU by 1.34\% and 1.46\%, respectively, and reducing the HD95 by 1.34 mm. Compared to boundary detection task, boundary-guided feature enhancement module brings larger performance improvement. This results validate the proposed BGFE could completely utilize the prior information from boundary to enhance the feature representations, thus improving the segmentation performance. Moreover, the SAAM module also brings a slight improvement over the baseline, with an increase of 0.30\% in Dice and 0.44\% in IoU, a reduction of 0.32 mm in HD95. This result could demonstrate the effectiveness of the SAAM module. Finally, the combination of BD, BGFE, and SAAM achieves the best overall performance, as Dice of 85.34\%, IoU of 77.52\%, and HD95 of 11.66 mm. Compared to the baseline, the Dice and IoU increases by 2.29\% and 2.61\%, respectively, while the HD95 decreasing by 2.68 mm. These results demonstrate the effectiveness and complementary of the proposed BD, BGFE, and SAAM. The BD and BGFE could guide the model focus on the boundary regions, and the SAAM module could adaptively adjust the varying sizes of tumor regions.

Fig.~\ref{fig:FigAblation} visualizes the segmentation results in the ablation experiment. As illustrated in Fig.~\ref{fig:FigAblation}, the baseline model(B) struggles when dealing with low-contrast regions or fuzzy boundaries, and often produces under-segmentation or jagged edges. These specific erroneous regions are highlighted by red boxes. The introduction of the boundary detection(BD) module improves the ability of the model to delineate lesion contours. Furthermore, the integration of the boundary-guided feature enhancement (BGFE) module, and the scale-aware aggregation module (SAAM) significantly enhances internal compactness and morphological consistency. Ultimately, the complete model integrating all three modules demonstrates superior segmentation performance. It effectively rectifies the defective regions observed in the baseline, and generates prediction masks with smooth boundaries, and clear details which exhibit a high degree of consistency with the ground truth. 

\begin{figure}[!t]
    \centering
    \includegraphics[width=\linewidth]{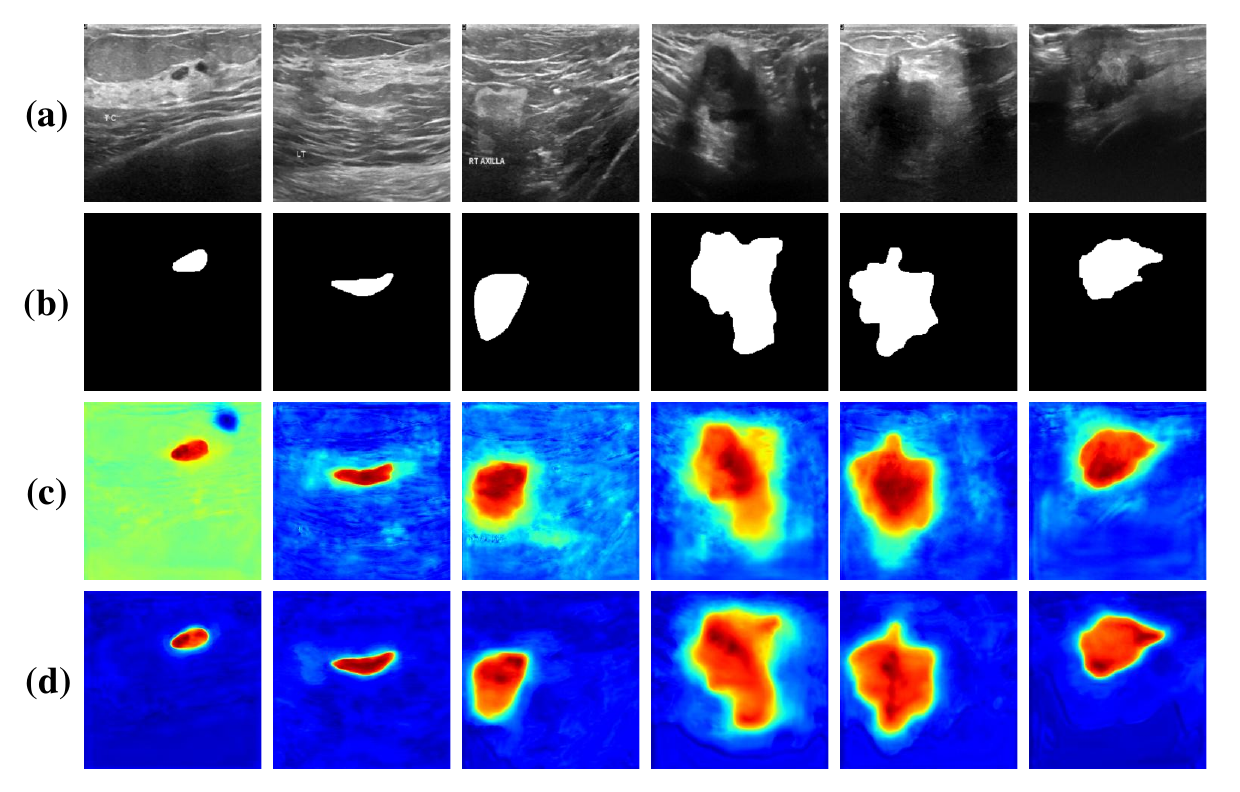}
    \caption{Some comparison examples of  baseline model and our proposed method. (a) original image, (b) GT, (c) heat map w/o BGFE, (d) heat map with BGFE. From the above visualization results, we can find that the BGFE could effectively decrease the background noise and fucus on the lesion regions.}
    
    \label{fig:FigJet1}
\end{figure}

In addition, we visualize the heat maps of the baseline model and our proposed method in Fig.~\ref{fig:FigJet1}. It can be observed that the heat maps of the baseline model contain more background noise. In contrast, our method effectively focuses on the lesion regions and reduces background noise. These results could demonstrate the validity of our proposed BGFE module. The BGFE module could utilize the prior information from the boundary and enable the module to concentrate on the extraction and analysis of tumor-related features.

\begin{figure}[!t]
    \centering
    \includegraphics[width=\linewidth]{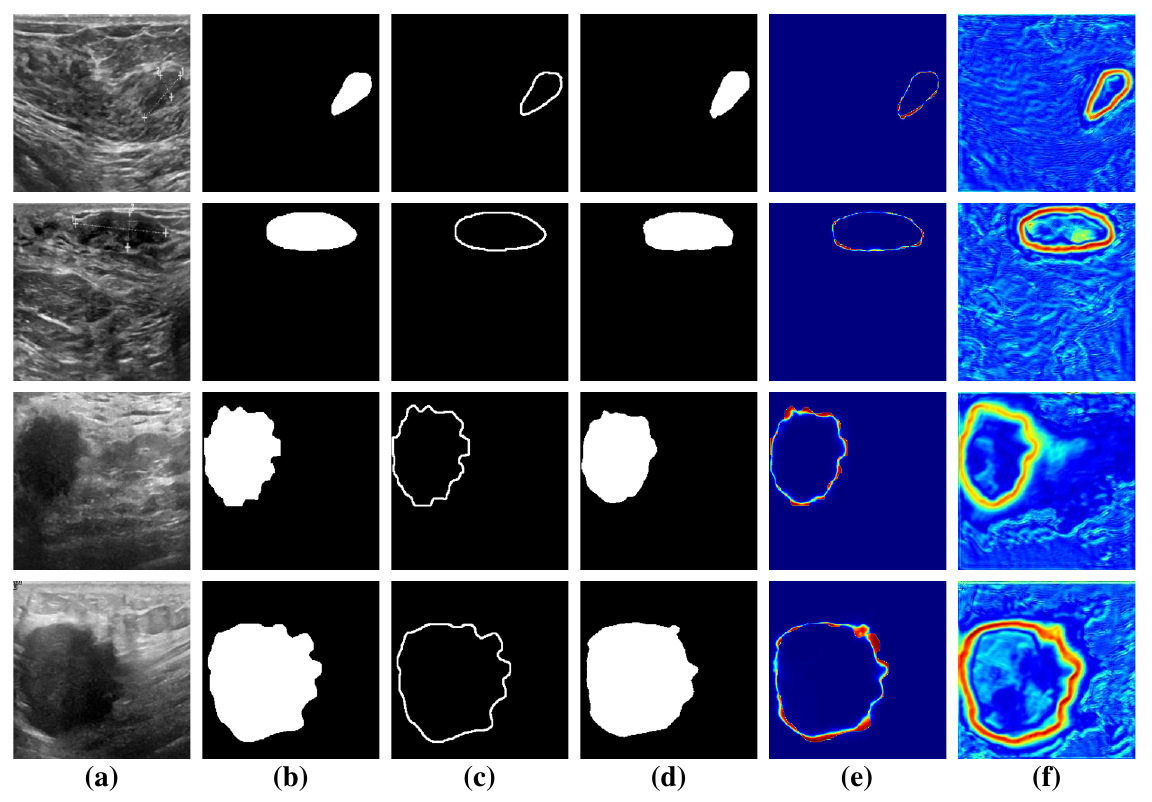}
    \caption{Visualization of boundary uncertainty and the proposed boundary attention maps. (a) original image, (b) GT, (c) GT boundary, (d) prediction map of baseline model, (e) error map, (f) boundary attention map generated by our method.}
    \label{fig:FigBoundaryAttention}
\end{figure}
Additionally, Fig.~\ref{fig:FigBoundaryAttention} demonstrates the effectiveness of the proposed Boundary-guided Feature Enhancement(BGFE) module. Baseline models often exhibit significant errors at the boundaries of target regions. These errors are clearly represented in the uncertainty maps. High uncertainty levels appear primarily along the edges of the segmented objects. The proposed BGFE module generates specialized boundary attention maps to address this issue. These attention maps show high activation in the exact regions where the baseline model remains uncertain. This alignment indicates that the boundary attention mechanism successfully directs the network to focus on challenging edge features. By highlighting these critical areas the module helps the model reduce ambiguity and improve segmentation precision. The visualization confirms that the BGFE module provides strong guidance for boundary refinement.

\begin{figure}[!t]
    \centering
    \includegraphics[width=\linewidth]{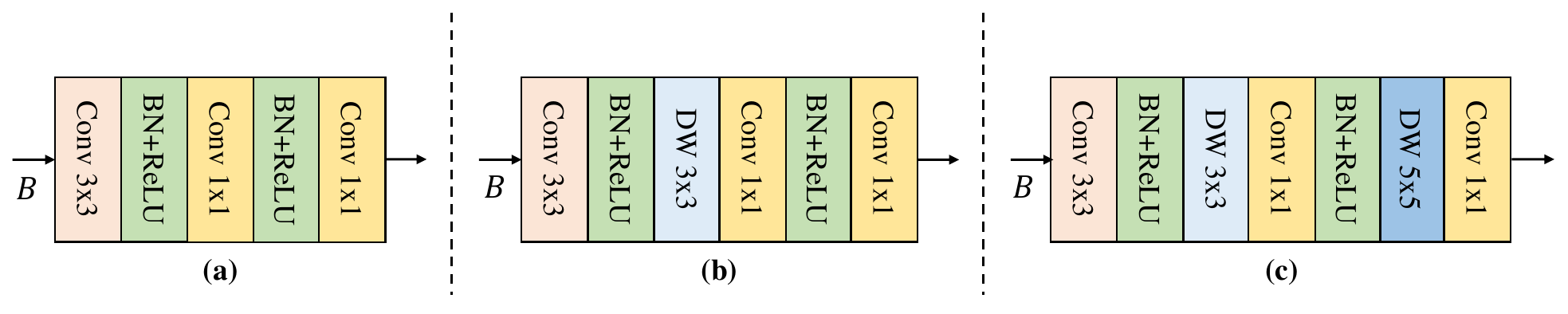}
    \caption{Different operations of boundary in BGFE. (a) refers to \textit{w/o} DW $3\times3$ and $5\times5$ , (b) refers to \textit{w/.} DW $3\times3$, (c) refers to \textit{w/.} DW $3\times3$ and $5\times5$, which is used in BGFE.}
    \label{fig:FigAblationOfBGFE}
\end{figure}

\begin{table}[!t]
\label{tab:tabAblationOfBGFE}
\caption{Results of Ablation study of different operations for boundary on the BUSI dataset. Here ``DW $3\times3$'' and ``DW $5\times5$'' mean the Depth-wise convolutions with kernel size as $3\times3$ and $5\times5$. }
\centering
\resizebox{1\linewidth}{!}{
\begin{tabular}{l|cc|ccccc}
\toprule
 &DW $3\times3$   &DW $5\times5$ &Dice$\%\color{red}{\uparrow}$&IoU$\%\color{red}{\uparrow}$&HD95$\color{red}{\downarrow}$&Recall$\%\color{red}{\uparrow}$&Acc$\%\color{red}{\uparrow}$ \\
\midrule
(a) & $\times$ & $\times$ &83.59 &75.41 &14.95 &85.83 &96.50 \\
(b) & $\checkmark$ & $\times$ &84.47 &76.63 &13.05 &85.82 &96.96 \\
(c) & $\checkmark$ & $\checkmark$ &\textbf{85.34} &\textbf{77.52} &\textbf{11.66} &\textbf{87.95} &\textbf{97.16}\\
\bottomrule
\end{tabular}}
\end{table}

\subsubsection{Effectiveness of Multi-Scale Depth-wise Convolutions in the BGFE Module}

In this section, we conduct experiments to evaluate the impact of multi-scale depth-wise convolutions for expanding the boundary region within the BGFE module. Fig.~\ref{fig:FigAblationOfBGFE} shows the comparison of different operational designs and Table~\ref{tab:tabAblationOfBGFE} presents the quantitative results. As shown in Table~\ref{tab:tabAblationOfBGFE}, both the depth-wise $3\times3$ and $5\times5$ convolutions contribute to performance gains. The depth-wise $3\times3$ convolution alone brings the Dice to 84.47\%, while integrating the depth-wise $5\times5$ further boosts the Dice to 85.34\%. These results suggest that the large kernel depth-wise convolutions effectively expand the influence areas of the boundary, thereby improving the segmentation performance.

\subsubsection{Effectiveness of different feature enhancement methods}

In this section, we compare different boundary-guided feature enhancement strategies on the BUSI dataset: Addition, Element-wise Multiplication, Channel Concatenation, and our proposed BGFE. 
As shown in Table~\ref{tab:FEofBGFE}, our proposed BGFE module markedly outperforms all alternative fusion strategies, achieving a Dice score of 85.34\%, IoU of 77.52\%, HD95 of 11.66 mm, Recall of 87.95\%, and Accuracy of 97.16\%. We guess that the Concatenation operation may introduce additional noise, while Addition and Multiplication may only affect a very narrow band along the boundary itself, limiting their effectiveness. 

To overcome these limitations and better exploit boundary information, we design the BGFE module. Instead of directly multiplying the raw boundary map with features, BGFE first enlarges the spatial influence of the boundary via multi‑scale depth‑wise convolutions with large kernels. The resulting expanded attention map is then used to modulate the feature maps. As a result, BGFE can cover a broader region around the boundary where segmentation uncertainty typically occurs, leading to more accurate and robust segmentation.



\begin{table}[!t]
\caption{Comparison with various feature enhancement methods on the BUSI dataset.}
\centering
\resizebox{\linewidth}{!}{
\begin{tabular}{lccccc}
\toprule
 & Dice$\%\color{red}{\uparrow}$&IoU$\%\color{red}{\uparrow}$&HD95$\color{red}{\downarrow}$&Recall$\%\color{red}{\uparrow}$&Acc$\%\color{red}{\uparrow}$ \\
\midrule
Addition &84.30 &76.09 &14.11 &86.16 &96.61 \\
Multiply &84.62 &76.71 &12.97 &86.89 &96.63 \\
Concatenate &84.83 &76.84 &12.74 &87.14 &96.79 \\

BGFE (Ours) &\textbf{85.34} &\textbf{77.52} &\textbf{11.66} &\textbf{87.95} &\textbf{97.16}\\

\bottomrule
\label{tab:FEofBGFE}
\end{tabular}}
\end{table}

\begin{table}[!t]
\caption{Performance comparison of integrating the BGFE at different network stages on the BUSI dataset. Here we employ the boundary detection module in the ``Encoder'', ``Decoder'', and both stages.}
\centering
\resizebox{\linewidth}{!}{
\begin{tabular}{ccccccc}
\toprule
Encoder & Decoder& Dice$\%\color{red}{\uparrow}$&IoU$\%\color{red}{\uparrow}$&HD95$\color{red}{\downarrow}$&Recall$\%\color{red}{\uparrow}$&Acc$\%\color{red}{\uparrow}$ \\
\midrule
$\times$& $\times$ &83.35 &75.35 &14.02 &84.24 &97.07 \\


\checkmark &  $\times$ & 83.53 &75.63 &13.97 &85.09 & 97.10\\
$\times$&\checkmark  &\textbf{85.34} &\textbf{77.52} &\textbf{11.66} &\textbf{87.95} &\textbf{97.16}\\
\checkmark & \checkmark &84.59 &76.50 &12.86 &86.72 &97.13\\
\bottomrule
\label{tab:tabBGFEStage}
\end{tabular}}
\end{table}
\subsubsection{Effectiveness of different stages for the boundary-guided feature enhancement module}

To investigate the impact of our boundary-guided feature enhancement (BGFE) module at different network stages, we conduct experiments by placing BGFE in encoder-only, encoder+decoder, and decoder-only. As shown in Table.~\ref{tab:tabBGFEStage}, we can find that the BGFE module could always brings performance improvements whether it is placed in the encoder or the decoder stage. In addition, integrating the BGFE module in the decoder stage achieved the best performance with Dice of 85.34\%, IoU of 77.52\%, and HD95 of 11.66 mm, while encoder-only yields the worst performance. Compared with the decoder, placing the boundary module in both the encoder and the decoder leads to a significant performance drop. We hypothesize the performance drops may be related to the distinct attributes of features at different stages. Progressive down-samplings in the encoder degrade the boundary information and limit the effectiveness of the BGFE, while up-samplings in the decoder enriching the boundary details and maximizing its feature refinement capability.

\subsubsection{Effectiveness of Varying Loss weight $\lambda_2$ for Boundary Detection}
\label{subsubsection:lossweight}
To determine the affect of different values of $\lambda_2$, we conducted experiments on the BUSI dataset. We set $\lambda_2$ as 0.0, 0.1, 0.3, 0.5, 0.7 and 0.9, respectively. It is notable that $\lambda_2=0$ indicate the predicted boundary is not supervised by the boundary GT. And the BGFE module is reduced to a spatial attention module. As can be seen in Table.~\ref{tab:tabVariousBoundaryLossWeight}, even when $\lambda_2=0$, our BGFE still slightly surpasses the baseline module with the SAAM module. As the weight $\lambda_2$ increases, the main metrics first improve and then slightly drop. Our method achieves the optimal performance at $\lambda_2=0.7$, which indicates the critical balance between boundary detection and primary tumor segmentation.

\begin{table}[!t]
\caption{Ablation Study of the Loss Weight $\lambda_2$ on the BUSI Validation Dataset. We set the $\lambda_2$ as $0, 0.1,0.3, 0.5, 0.7,0.9$.  }
\centering
\resizebox{\linewidth}{!}{
\begin{tabular}{lccccc}
\toprule
&Dice$\%\color{red}{\uparrow}$&IoU$\%\color{red}{\uparrow}$&HD95$\color{red}{\downarrow}$&Recall$\%\color{red}{\uparrow}$&Acc$\%\color{red}{\uparrow}$ \\
\midrule
$\lambda_2 = 0.0$ &83.97 &75.96 &13.99 &86.27 &96.60\\
$\lambda_2 = 0.1$ &84.35 &76.21 &13.44 &86.86 &96.66 \\
$\lambda_2 = 0.3$ &84.48 &76.33 &13.02 &87.37 &96.52 \\
$\lambda_2 = 0.5$ &85.13 &77.02 &12.54 &87.28 &96.93 \\
$\lambda_2 = 0.7$ &\textbf{85.34} &\textbf{77.52} &\textbf{11.66} &\textbf{87.95} &\textbf{97.16}\\
$\lambda_2 = 0.9$ &84.68 &76.55 &13.09 &87.31 &96.67 \\
\bottomrule
\label{tab:tabVariousBoundaryLossWeight}
\end{tabular}}
\end{table}



\subsection{ Cross-architecture Generalization of the BGFE Module}
To further validate the generalizability of the proposed Boundary-Guided Feature Enhancement (BGFE) module, we integrate it into four distinct baseline architectures: U-Net~\cite{ronneberger2015u}, TransUNet~\cite{chen2021transunet}, CMUNet~\cite{tang2023cmu}, and CMUNeXt~\cite{tang2024cmunext}. As summarized in Table~\ref{tab:BGFEinOtherModels}, the BGFE module could consistently improve the segmentation performance across all baselines with only a marginal increase in parameters and GFLOPs. Notably, it achieves the most substantial improvement for U-Net, elevating the Dice by 8.53\% and IoU by 8.93\%, while reducing HD95 by 9.78 mm. Even for the SOTA breast tumor segmentation method, CMUNeXt, our BGFE module still brings $1.60\%$ and $1.80\%$ improvements for Dice and IoU metrics. These results demonstrate that the BGFE module is not architecture-specific and possesses strong generalization capability, effectively enhancing segmentation performance across diverse network designs.


\begin{table}[!t]
\caption{Cross-architecture evaluation of the BGFE module on the BUSI dataset. The proposed module yields consistent performance improvements across all baseline models.}
\centering
\resizebox{\linewidth}{!}{
\begin{tabular}{llllll}
\toprule
 &Params(M)$\color{red}{\downarrow}$ &GFLOPs$\color{red}{\downarrow}$ &Dice$\%\color{red}{\uparrow}$ &IoU$\%\color{red}{\uparrow}$ &HD95$\color{red}{\downarrow}$\\
\midrule
U-Net &34.53 &65.52 &75.57 &67.38 &23.61 \\
U-Net + BGFE&42.51$\color{red}({\uparrow}8.01)$ &90.17$\color{red}({\uparrow}24.65)$ &84.10$\color{red}({\uparrow}8.53)$ &76.31$\color{red}({\uparrow}8.93)$ &13.83$\color{red}({\downarrow}9.78)$ \\
\midrule
TransUNet &93.23 &32.23 &82.94 &74.79 &16.37 \\
TransUNet + BGFE&95.21$\color{red}({\uparrow}1.98)$ &37.26$\color{red}({\uparrow}5.03)$ &83.88$\color{red}({\uparrow}0.94)$ &75.80$\color{red}({\uparrow}1.01)$ &14.45$\color{red}({\downarrow}1.92)$ \\
\midrule
CMUNet &49.93 &91.25 &82.62 &74.64 &14.79\\
CMUNet + BGFE&57.91$\color{red}({\uparrow}7.98)$ &115.87$\color{red}({\uparrow}24.62)$ &83.93$\color{red}({\uparrow}1.31)$&75.81$\color{red}({\uparrow}1.17)$ &13.23$\color{red}({\downarrow}1.56)$ \\
\midrule
CMUNeXt &3.15 &7.42 &83.05 &74.91 &14.34\\
CMUNeXt + BGFE&4.15$\color{red}({\uparrow}1.00)$ &10.36$\color{red}({\uparrow}2.94)$ &84.65$\color{red}({\uparrow}1.60)$ &76.71$\color{red}({\uparrow}1.80)$ &12.91$\color{red}{\downarrow}(1.43)$ \\
\bottomrule
\label{tab:BGFEinOtherModels}
\end{tabular}}
\end{table}

\begin{table}[!t]
\caption{Robustness analysis on the BUSI and Dataset B. BUSI$\rightarrow$Dataset B means the strategy of training on the BUSI dataset and validation on the Dataset B. }
\label{tab:RobustnessAndGeneralization1}
\centering
\resizebox{\linewidth}{!}{
\begin{tabular}{l|cc|cc}
\toprule
\multirow{2}{*}{Methods}& \multicolumn{2}{c|}{BUSI$\rightarrow$BUSI} & \multicolumn{2}{c|}{Dataset B$\rightarrow$BUSI}  \\
\cline{2-5}
 & Dice$\%\color{red}{\uparrow}$ 
 & IoU$\%\color{red}{\uparrow}$
 & Dice$\%\color{red}{\uparrow}$ 
 & IoU$\%\color{red}{\uparrow}$\\
\midrule
U-Net (MICCAI2015)~\cite{ronneberger2015u} &75.57 &67.38 &57.50 &47.59   \\
TransUNet (ARXIV2021)~\cite{chen2021transunet} &82.94 &74.79 &58.20 &47.95 \\
GGNet (MIA2021)~\cite{xue2021global} &82.63 &74.23 &64.26 &54.86\\
BGNet (IJCAI2022)~\cite{sun2022bgnet} &83.05 &74.98 &54.79 &45.21 \\
CMUNet (ISBI2023)~\cite{tang2023cmu} &82.62 &74.64 &56.99 &47.80  \\
CMUNeXt (ISBI2024)~\cite{tang2024cmunext} &83.05 &74.91 &53.94 &45.89 \\
BFNet (TCSVT2025)~\cite{yue2025boundary}&81.45 &73.32  &63.18  &54.25 \\
PBE-UNet(ours) &\textbf{85.34} &\textbf{77.52} &\textbf{66.31} &\textbf{56.04}\\
\midrule

\multirow{2}{*}{Methods} & \multicolumn{2}{c|}{Dataset B$\rightarrow$Dataset B} & \multicolumn{2}{c}{BUSI$\rightarrow$Dataset B} \\
\cline{2-5}
 & Dice$\%\color{red}{\uparrow}$ 
 & IoU$\%\color{red}{\uparrow}$
 & Dice$\%\color{red}{\uparrow}$ 
 & IoU$\%\color{red}{\uparrow}$\\
\midrule
U-Net (MICCAI2015)~\cite{ronneberger2015u} &80.29 &72.90  &69.94  &61.67  \\
TransUNet (ARXIV2021)~\cite{chen2021transunet} &83.77 &75.28  &76.39  &68.22  \\
GGNet (MIA2021)~\cite{xue2021global} &84.80 &76.04  &\textbf{81.47}  &\textbf{72.36} \\
BGNet (IJCAI2022)~\cite{sun2022bgnet} &87.63 &79.65  &69.98  &59.83 \\
CMUNet (ISBI2023)~\cite{tang2023cmu} &84.20 &75.78  &76.29  &67.42 \\
CMUNeXt (ISBI2024)~\cite{tang2024cmunext} &82.85 &74.14  &77.02  &68.86 \\
BFNet (TCSVT2025)~\cite{yue2025boundary} &82.79 &73.89  &80.17  &71.16 \\
PBE-UNet(ours) &\textbf{88.50} &\textbf{81.02}  &80.22  &71.72 \\

\bottomrule
\end{tabular}}
\end{table}
\subsection{Robustness Analysis}

To evaluate the robustness of the model, we conducted experiments on various models under both in-domain and cross-domain scenarios. As shown in Tab.~\ref{tab:RobustnessAndGeneralization1}, our proposed PBE-UNet achieves the best Dice and IoU in three out of four experimental settings. In in-domain testing, our model yielded state-of-the-art performance across all datasets. On the BUSI dataset, PBE-UNet achieves Dice of 85.34\% and IoU of 77.52\%. On the Dataset B dataset, it attains Dice of 88.50\% and IoU of 81.02\%. These results confirm the robust inherent learning capability of our model. In cross-domain testing, when trained on the Dataset B dataset and tested on the BUSI dataset, our model significantly outperforms all compared models with 66.31\% Dice and 56.04\% IoU. In the other cross-domain setting, trained on the BUSI dataset and tested on the Dataset B dataset, the GGNet model achieves the highest scores of 81.47\% Dice and 72.36\% IoU, while our model follows with 80.22\% Dice and 71.72\% IoU. It is noteworthy that some models exhibit severe performance degradation in cross-domain scenarios, whereas our model maintains relatively high and stable performance across all cross-domain settings. These findings collectively demonstrate that PBE-UNet not only excels in in-domain segmentation tasks, but also acquires more intrinsic feature representations robust to data distribution shifts resulting in superior, and well-balanced overall generalization performance.

\subsection{Parameters and FLOPs Analysis}

We further assessed the proposed method in terms of computational complexity, which primarily involves the number of parameters and FLOPs. The proposed PBE-UNet demonstrates notably low computational demands, with 4.26 million parameters and 10.72 GFLOPs. This places it among the most efficient models listed, significantly more lightweight than earlier benchmarks like U-Net or Att U-Net and considerably more efficient than larger contemporaries such as TransUNet, AAU-Net, or ESKNet. Notably, CMUNeXt and BFNet yield yet lower overheads per respective metric, with CMUNeXt exhibiting the fewest parameters and BFNet the lowest FLOPs. These results on computational complexity reflect the overall trend in recent years towards designing more efficient architectures, highlighting the emphasis on efficiency, especially in resource-constrained environments. The results suggest that PBE-UNet successfully balances a compact model size with manageable computational cost, making it a competitive option where efficiency is prioritized alongside predictive performance.

\begin{table}[!t]
\centering
\caption{The results of parameters and FLOPs comparison for different methods. We employ the ``Parameters'' and ``FLOPS'' to evaluate the complexity of different methods.}
\footnotesize
\resizebox{\linewidth}{!}{
\begin{tabular}{llll}
\toprule
Model & Pub' Year &Parameters (M)$\color{red}{\downarrow}$ & FLOPs (G)$\color{red}{\downarrow}$  \\ 
\midrule
U-Net& MICCAI2015~\cite{ronneberger2015u} & 34.53 & 65.59 \\
Att U-Net& MIDL2018~\cite{oktay2018attention} & 34.88 & 66.70 \\
GGNet & MIA2021~\cite{xue2021global} &59.47 &18.40 \\
TransUnet& Arxiv2021~\cite{chen2021transunet}& 93.20 & 24.70 \\
BGNet & IJCAI2022~\cite{sun2022bgnet} &77.80 &22.16 \\
AAU-Net&TMI2023~\cite{chen2022aau} & 158.51 & 249.37 \\
CMUNet&ISBI2023~\cite{tang2023cmu} & 49.93 & 91.34 \\
ESKNet&ESWA2024~\cite{chen2024esknet} & 114.82 & 196.37 \\
CMUNeXt&ISBI2024~\cite{tang2024cmunext}& 3.15 & 7.42 \\
PConv-UNet&DISPLAYS2025~\cite{wang2025pconv}  & 57.26 & 76.90 \\
BFNet &TCSVT2025~\cite{yue2025boundary} &25.49  &5.88 \\
PBE-UNet& Ours &4.26 &10.72\\
\bottomrule
\end{tabular}}
\label{tab:flops}
\end{table}

\subsection{Failure Cases}

\begin{figure}[!t]
    \centering
    \includegraphics[width=1\linewidth]{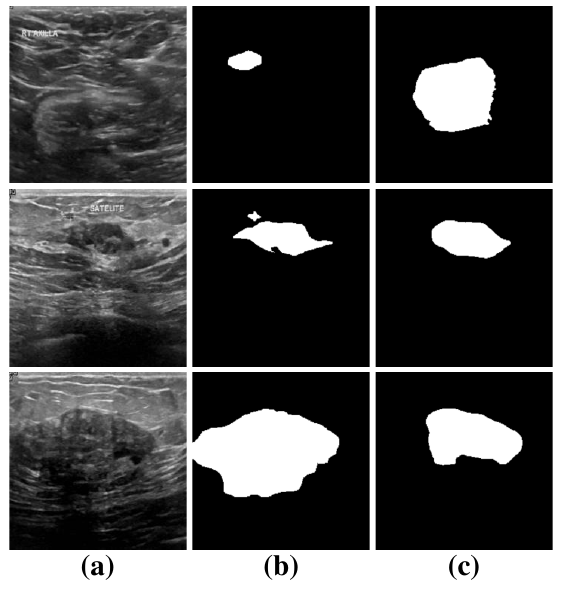}
    \caption{Some failure cases of PBE-UNet on the BUSI dataset. (a) original image, (b) GT, (c) ours. }
    \label{fig:faliureCases}
\end{figure}

Although the proposed PBE‑UNet delivers strong overall performance, its segmentation accuracy still declines in certain challenging cases. As shown in the first row of Fig.~\ref{fig:faliureCases}, the model exhibits severe over‑segmentation, with predicted regions substantially exceeding the ground‑truth annotations. This behavior mainly arises from strong background interference in the input images, where the network struggles to distinguish the target from complex surrounding textures. In the second and third rows, a pronounced under‑segmentation tendency is observed: the predicted boundaries are overly simplified, leading to the loss of fine‑grained morphological details. This issue is especially evident in regions with blurred boundaries and low contrast, where the model often fails to capture sufficient discriminative features. These results indicate that the algorithm remains challenged in processing edge regions with low signal‑to‑noise ratios. Overall, the failure cases suggest that the perception of local details in our PBE‑UNet requires further improvement, providing a clear direction for future architectural refinement.

\section{Conclusion}

In this work, we present PBE-UNet, a novel progressive boundary-enhanced framework for ultrasound image segmentation. First, we propose a light weight Scale-Aware Aggregation Module (SAAM) to tackle the challenge of scale variations for lesions. This module leverages parallel depthwise dilated convolutions with varying receptive fields to adaptively captures robust multi-scale feature representations. Then, we introduce the Boundary-Guided Feature Enhancement (BGFE) module to address the mismatch between the narrow boundaries and wide segmentation errors. This is achieved by expanding the narrow boundary prediction into broader spatial attention maps with large kernel convolutions. The expanded attention maps could effectively cover the wider segmentation error regions. Experiment results on four benchmark datasets, BUSI, Dataset B, TN3K, and BP, demonstrate the validity of our PBE-UNet in accurate ultrasound image segmentation. Ablation studies also confirm each module's contribution. The proposed spatial-adaptive boundary expansion paradigm, which transforms thin boundary priors into semantically wider attention regions, offers a new perspective for tracking boundary ambiguity in medical image segmentation.

\section*{References}
\bibliographystyle{splncs04}
\bibliography{Reference}

\end{document}